\documentclass{article}
\usepackage{arxiv}

\usepackage{epstopdf}
\usepackage{multirow} 
\usepackage{amsfonts}
\usepackage{caption}
\usepackage{subcaption}
\newcommand{\mli}[1]{\mathit{#1}}

\def \method{SALVE}
\def \natle{NATLE}

\usepackage{float}
\usepackage[table,xcdraw]{xcolor}
\usepackage{amsmath,amssymb}
\usepackage{graphicx}
\usepackage[utf8]{inputenc} 
\usepackage[T1]{fontenc}    
\usepackage{hyperref}       
\usepackage{url}            
\usepackage{booktabs}       
\usepackage{amsfonts}       
\usepackage{nicefrac}       
\usepackage{microtype}      
\usepackage{lipsum}
\DeclareMathOperator*{\argmin}{argmin}

\title{\method{}: Self-Supervised Adaptive Low-Light Video Enhancement}

\author{
  Zohreh Azizi \\
  Media Communications Lab\\
  University of Southern California\\
  Los Angeles, CA, USA \\
  \texttt{zazizi@usc.edu} \\
   \And
  C.-C. Jay Kuo \\
  Media Communications Lab\\
  University of Southern California\\
  Los Angeles, CA, USA \\
  \texttt{cckuo@sipi.usc.edu} \\
}

\begin{document}
\maketitle

\begin{abstract}
A self-supervised adaptive low-light video enhancement method, called
\method{}, is proposed in this work. \method{} first enhances
a few keyframes of an input low-light video using a retinex-based
low-light image enhancement technique. For each keyframe, it learns a
mapping from low-light image patches to enhanced ones via ridge
regression. These mappings are then used to enhance the remaining frames
in the low-light video. The combination of traditional retinex-based
image enhancement and learning-based ridge regression leads to a robust,
adaptive and computationally inexpensive solution to enhance low-light
videos. Our extensive experiments along with a user study show that 87\%
of participants prefer \method{} over prior work.
\end{abstract}


\section{Introduction}\label{sec:intro}

Videos captured under low light conditions are often noisy and of poor
visibility.  Low-light video enhancement aims to improve viewers'
experience by increasing brightness, suppressing noise, and amplifying
detailed texture.  The performance of computer vision tasks such as
object tracking and face recognition can be severely affected under
low-light noisy environments.  Hence, low-light video enhancement is
needed to ensure the robustness of computer vision systems. Besides, the
technology is highly demanded in consumer electronics such as video
capturing by smart phones.  

While mature methods for low-light \textit{image} enhancement have been
developed in recent years, low light \textit{video} enhancement is still
a standing challenge and open for further improvement.  A trivial
solution to low light video enhancement is to enhance each frame with an
image enhancement method independently.  However, since this solution
disregards temporal consistency, it tends to result in flickering videos
\cite{lai2018learning}.  Also, frame-by-frame low light video processing 
can be too computationally expensive for practical applications.

Several methods utilized deep learning (DL) to preserve the temporal
consistency of video frames. For instance, 3D CNNs are
trained to process a number of frames simultaneously in order to take
temporal consistency into account \cite{lv2018mbllen,
jiang2019learning}. Some papers enforce similarity between pairs of
frames with a temporal loss function or loss function regularization in
training \cite{chen2019seeing, eilertsen2019single}. Others works
extract the motion information and leverage redundancy among frames to
ensure the temporal consistency of enhanced videos
\cite{lai2018learning, zhang2021learning}. 

On one hand, the efforts mentioned above lead to high-performance models
with a range of acceptable to excellent quality results. On the other
hand, their performance is dependent on the training dataset.
Differences between the training and testing environments can degrade
the performance of low light video enhancement severely. In other
words, when deployed in the real world, the DL-based models cannot be
trusted and utilized without fine-tuning.  Considering the fact that
paired low-light/normal-light video datasets are very scarce,
fine-tuning these models can be challenging. 

In this paper, we propose an alternative low-light video enhancement
method to address the above-mentioned challenges.  Our proposed method
is called the self-supervised adaptive low-light video enhancement
(\method{}) method. By self-supervision, we mean that \method{} directly
learns to enhance an arbitrary input video without requiring to be
trained on other training videos. 

\method{} offers a robust solution that is highly adaptive to new
real-world conditions. \method{} selects a couple of keyframes from the
input video and enhances them using an effective retinex-based image
enhancement method called \natle{} \cite{azizi2020noise}. Given
\natle{}-enhanced keyframes of the input video, \method{} learns a
mapping from low-light frames to enhanced ones via ridge regression.
Finally, \method{} uses this mapping to enhance the remaining frames.
\method{} does not need low- and normal-light paired videos in training.
Therefore, it can be an attractive choice for non-public environments
such as warehouses and diversified environments captured by phone
cameras. 

\method{} is a hybrid method that combines components from a
retinex-based image enhancement method and a learning-based method. The
former component leads to a robust solution which is highly adaptive to
new real-world environments. The latter component offers a fast,
computationally inexpensive and temporally consistent solution. We
conduct extensive experiments to show the superior performance of
\method{}. Our user study shows that 87\% of participants prefer
\method{} over prior work. 

The rest of this paper is organized as follows. Related work is
discussed in Section~\ref{sec:related}. The low light image enhancement
method, \natle{}, is reviewed and then the proposed low light video
enhancement method, \method{}, is explained in Section~\ref{sec:method}.
Experimental results are presented in Section \ref{sec:experiments}.
Finally, concluding remarks are given in Section~\ref{sec:conclusion}. 

\section{Related Work}\label{sec:related}

\subsection{Low Light Image Enhancement}

There are two categories of traditional low-light image enhancement
methods: histogram equalization and retinex decomposition. Histogram
equalization stretches the color histogram to increase the image
contrast. Although it is simple and fast, it often yields unnatural
colors, amplifies noise, and under/over-exposes areas inside an image.
To address these artifacts, more complex priors are adopted for
histogram-based image enhancement~\cite{arici2009histogram,
celik2011contextual, ibrahim2007brightness, lee2013contrast,
nakai2013color}. Specific penalty terms were designed and used to
control the level of contrast enhancement, noise, and mean brightness in
\cite{arici2009histogram}. Inter-pixel contextual information was used
for non-linear data mapping in \cite{celik2011contextual}.  To preserve
the mean brightness, histogram equalization was applied to different
dynamic ranges of a smoothed image in \cite{ibrahim2007brightness}. The
gray-level differences between adjacent pixels were amplified to enhance
image contrast based on layered difference representation of 2D
histograms in \cite{lee2013contrast}.  Differential gray-level histogram
equalization was proposed in \cite{nakai2013color} based on the concept
of differential histograms. 

Inspired by the human vision system (HVS), it is assumed in the retinex
theory \cite{land1977retinex} that each image can be decomposed into two
components: a reflectance (R) term containing inherent properties and an
illumination (L) term containing the lightness condition. Along this
line, another approach for low light image enhancement is to decompose
an input image into R and L terms and adjust the L term to the
normal-light condition. Earlier work focused on R and L decomposition
and attempted to acquire R and L more
accurately~\cite{jobson1997properties, jobson1997multiscale} using a
Single-Scale Retinex (SSR) representation. Later, SSR was extended to a
MultiScale Retinex (MSR) representation, which can be used for color
image restoration.  An adaptive MSR was proposed in
\cite{lee2013adaptive}, which computes the weights of an SSR according
to the content of the input image. More recently, optimization functions
were carefully designed in \cite{ren2020lr3m, azizi2020noise} to
determine the R and L terms. They attempted to find a balance in
suppressing noise and preserving texture through the optimization
functions. 

Recently, the deep-learning (DL) paradigm has been proposed for
low-light image enhancement based on retinex theory \cite{wei2018deep,
zhang2019kindling, wang2019rdgan}. A decomposition network and an
illumination network were trained to perform retinex decomposition and
enhancement, respectively, in \cite{wei2018deep}.  The work in
\cite{zhang2019kindling} added another network, called reflectance
restoration, to mitigate color distortion and noise.  A generative
adversarial network (GAN) \cite{goodfellow2014generative} was employed
in \cite{wang2019rdgan} to generate decomposed and enhanced images.
Another GAN work \cite{jiang2021enlightengan} was trained without paired
data. The application of auto-encoders to image enhancement was
investigated in~\cite{lore2017llnet}. A multi-branch network was
proposed in \cite{lv2018mbllen} to extract rich features in different
levels for enhancement via multiple subnets.  An end-to-end network for
raw camera image enhancement was proposed in \cite{chen2018learning}.

\subsection{Low Light Video Enhancement}

While low-light image enhancement is a well-studied topic,
low-light video enhancement is still an ongoing and challenging research
topic. Applying image-based algorithms to each frame of a video yields
flickering artifacts due to inconsistent enhancement results along
time~\cite{lai2018learning}.  It is essential to take both
temporal and spatial information into account in video processing. One
approach is to extend 2D convolutional neural networks (CNNs) to 3D CNNs
\cite{lv2018mbllen}, which includes the 2D spatial domain and the 1D
temporal domain. A 3D U-net \cite{ronneberger2015u} was proposed in
\cite{jiang2019learning} to enhance raw camera images. However, these 3D
DL-based methods have huge model sizes and extremely high computational
costs. 

Another approach is to exploit self-consistency~\cite{chen2019seeing,
eilertsen2019single}. The resulting methods operate on single frames of
video but impose the similarity constraint on image pairs to improve the
performance and stability of their models.  A new static video dataset
was proposed in \cite{chen2019seeing}, containing short- and
long-exposure images of the same scene.  They took two random frames
from the same sequence in training and utilized the self-consistency
temporal loss to make the network robust against noise and small changes
in the scene. Different motion types were accounted for by imposing
temporal stability using a regularized cost function in
\cite{eilertsen2019single}. Another family of self-consistency-based
methods \cite{lai2018learning, zhang2021learning} used the optical flow
to estimate the motion information in a sequence.  They utilized the
FlowNet \cite{ilg2017flownet} to predict the optical flow between two
frames, and warped the frames based on the predicted flow to avoid
inconsistent frame processing. An image segmentation network was
exploited to detect the moving object regions before optical flow
prediction in \cite{zhang2021learning}.  A model to reduce noise and
estimate illumination was proposed in \cite{wang2021seeing} based on the
retinex theory. It took each frame along with two past and two future
frames as input to enhance the middle frame. 

All existing methods on low-light video enhancement employ deep neural
networks (DNNs) as their backbone. In this work, we propose an effective
and high performance method called \method{} to achieve the same goal
without the use of DNNs. Our method contributes to green video
processing with a lower carbon footprint \cite{kuo2023green,
azizi2022pager, rouhsedaghat2021successive}. Additionally, \method{}
does not need a training dataset; it is a self-supervised approach
which utilizes the frames of the test video and adapts its enhancement
strategy accordingly. As such, our approach does not rely on massive
training datasets and is robust against environmental changes.

\begin{figure*}[t]
\centering
\includegraphics[width=\linewidth]{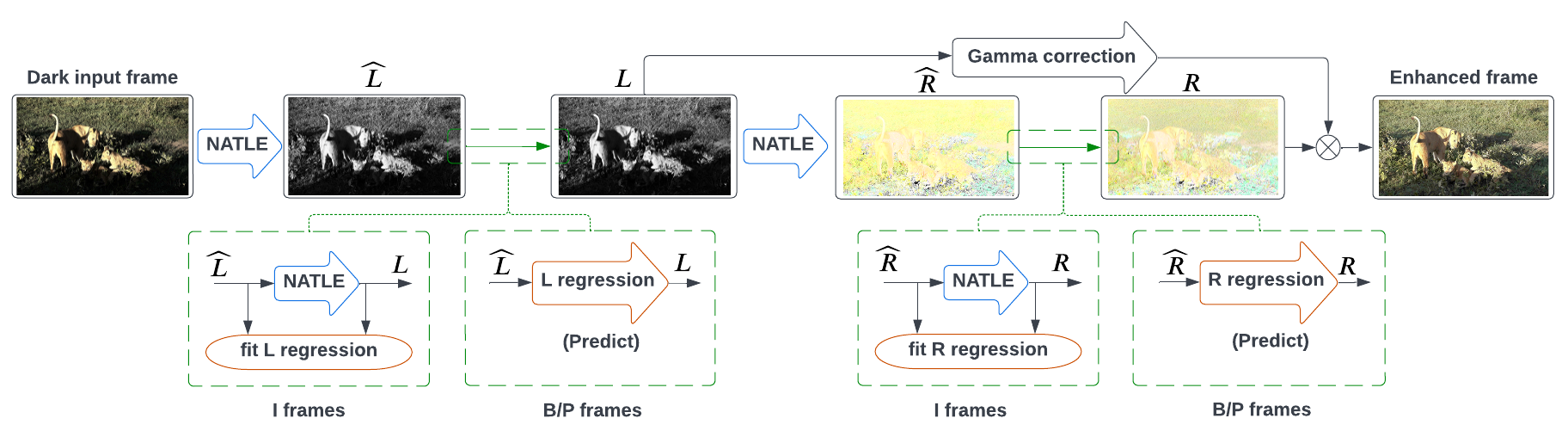}
\caption{An overview of the proposed \method{} method.  For intra-coded
frames (I frames), it estimates the illumination ($\widehat{L}$)
component and the reflectance ($\widehat{R}$) components using the NATLE
method.  For inter-coded frames (P/B frames), it predicts these
components using a ridge regression learned from the last raw and
enhanced $I$ frame pairs.} \label{fig:overview}
\end{figure*}

\section{Proposed Method}\label{sec:method}

Figure \ref{fig:overview} presents an overview of our proposed method.
The top row shows the steps taken to enhance an input frame, which we
discuss in Section \ref{subsec:NATLE}. The bottom row shows the
extension to videos, i.e. it shows how we treat different frames of the
video. We discuss this process in Section \ref{subsec:Video}.

\subsection{\natle{}}\label{subsec:NATLE}

In order to propose our method in Section \ref{subsec:Video}, we need to
first review \natle{} \cite{azizi2020noise}, which is an effective
method for low light image enhancement. \natle{} is a retinex-based low
light image enhancement method. A classic retinex model decomposes an
input image $S$ into the element-wise multiplication of two components;
a reflectance map ($R$) and an illumination ($L$) map:
\begin{equation}\label{eq:retinex}
S = R \circ L,
\end{equation}
where $R$ represents the inherent features within an image which remain
the same in different lightness conditions. $L$ shows the lightning
condition. Ideally, $R$ contains all the texture and details within the
image and $L$ is a piece-wise smooth map with significant edges.
\natle{} presents a methodology to find solutions for $R$ and $L$. It
then enhances $L$ to a normal light condition, and follows the retinex
model to combine the enhanced $L$ with $R$ and obtain the enhanced
image.  In what follows, we explain the steps \natle{} takes to find
solutions for $R$ and $L$.

\textbf{Step 1.} Calculate an initial estimation of $L$, namely
$\widehat{L}$, which is a weighted average of RGB color channels:
\begin{equation}\label{eq:L_hat}
\widehat{L} = 0.299R+0.587G+0.114B.
\end{equation}

\textbf{Step 2.} Form an optimization function to find a piece-wise 
smooth solution for $L$ as
\begin{equation}\label{eq:L}
\argmin_L \|L-\widehat{L}\|_{F}^2+\alpha\|\nabla L\|_{1},
\end{equation}
where $\alpha$ is set to 0.015 and $\|\nabla L\|_{1}$ is approximated with
\begin{equation}\label{eq:Lappr}
\lim_{\epsilon \to 0^+}\sum_{x}\sum_{d\in\{h,v\}}\dfrac{(\nabla_{d}
L(x))^2}{\mid\nabla_{d}\widehat{L}(x)\mid+\epsilon}=\|\nabla L\|_{1},
\end{equation}
and where $d$ is the gradient direction and $v$ and $h$ indicate the
vertical and horizontal directions, respectively. Eq.(\ref{eq:L}) 
can be rewritten as
\begin{equation}\label{eq:L2}
\argmin_L \|L-\widehat{L}\|_{F}^2+\sum_{x}\sum_{d\in\{h,v\}} 
A_{d}(x)(\nabla_{d} L(x))^2,
\end{equation}
where
\begin{equation}\label{eq:Ad}
A_{d}(x)= \frac{\alpha}{\mid\nabla_{d}\widehat{L}(x)\mid+\epsilon}.
\end{equation}
Finally, Eq. (\ref{eq:L2}) is solved by differentiating with respect to
$L$ and setting the derivative to zero. The final solution is derived in
closed form as
\begin{equation}\label{eq:l}
l=(I+\sum_{d\in\{h,v\}}{D_{d}^T {\mbox Diag} (a_{d})D_{d}})^{-1}\widehat{l}, 
\end{equation}
where $\mbox{Diag} (a_d)$ is a matrix with $a_d$ on its diagonal,
$D_{d}$ is a discrete differential operator matrix that plays the role
of $\nabla$ and $I$ is the identity matrix. Once vector $l$ is
determined, it is reshaped to matrix $L$. 

\textbf{Step 3.} Calculate an estimate of $R$ in form of
\begin{equation}\label{eq:Rhat}
\widehat{R}=S\oslash(L+\varepsilon)-N,
\end{equation}
where $S$ is the input image, $L$ is the estimated illumination obtained
in Step 2, $-N$ shows a median filter denoising followed by a bilateral
filter denoising, $\oslash$ denotes element-wise division and
$\varepsilon$ is a small value to prevent division by zero. 

\textbf{Step 4.} Form an optimization function to find $R$ via
\begin{equation}\label{eq:R}
\argmin_R \|R-\widehat{R}\|_{F}^2+\beta\|\nabla R-G\|_{F}^2,   
\end{equation}
where $\beta$ is set to 3 in our experiments.  The first term in Eq.
(\ref{eq:R}) ensures that $R$ is noise-free and consistent with the
retinex model. The second term has a noise-removal and texture-preserving dual role. Then, we get
\begin{equation}\label{eq:G}
G=\begin{cases}
   0, \quad \quad  \nabla S<\epsilon_{g} \\
   \lambda\nabla S,
   \end{cases}
\end{equation}
where $\epsilon_{g}$ is the threshold to filter out small gradients,
which are viewed as noise, and $\lambda$ controls the degree of texture
amplification. The values of $\epsilon_{g}$ and $\lambda$ are set to
0.05 and 1.1, respectively. Finally, the optimization problem in Eq.
(\ref{eq:R}) is solved by differentiating with respect to $R$ and
setting the derivative to zero. The final solution can be derived in
closed form as
\begin{equation}\label{eq:r}
r=(I+\beta\sum_{d\in\{h,v\}}{D_{d}^2})^{-1}(\widehat{r}+ \beta \sum_{d\in\{h,v\}},
{D_{d}^T g_{d}}),
\end{equation}
where $D_{d}$, $\nabla$ and $I$ are defined the same as those in Eq.
(\ref{eq:l}). Once vector $r$ is determined, it is reformed to matrix
$R$. 

\textbf{Step 5.} Apply gamma correction to $L$ for illumination adjustment.
The ultimate enhanced image is computed as
\begin{equation}\label{eq:gamma}
S^{'} = R \circ L^{\frac{1}{\gamma}},
\end{equation}
where $\gamma$ is set to 2.2 in the experiment.

\subsection{Video Enhancement}\label{subsec:Video}

We showed the performance of \natle{} in low light image enhancement in
\cite{azizi2020noise}. \natle{} suppresses noise and preserves texture
while enhancing low-light images. In order to extend the application of
\natle{} to videos, the trivial idea would be to apply \natle{}
separately on all the frames within a video. However, a series of
consecutive frames within a video usually have significant correlations
in structure, color, and light. We may leverage this temporal similarity
in order to lower the costs of the video enhancement from a series of
repetitive image enhancements. 

Here, we propose a self-supervised method for low light video
enhancement based on learning from \natle{}. By applying \natle{} on
selected frames within a video, we acquire pairs of low-light and
enhanced frames, from which we learn a mapping from low-light to
enhanced frames. We then apply the learnt mapping to the rest of the
frames to accomplish the low light video enhancement. In
particular, we approximate Eqs. (\ref{eq:l}) and (\ref{eq:r}) in
\natle{}, which take the major portion of \natle{}\'s runtime.
Thus, our proposed video enhancement method is significantly faster
and computationally less expensive than applying frame-by-frame
\natle{}. 

In order to decide the frame on which we apply \natle{}, we use the
FFMPEG compression technique. In FFMPEG, there are three types of
frames, namely identity (\textit{I}), bidirectional (\textit{B}), and
predicted (\textit{P}) frames. The \textit{I} frames are the keyframes
which indicate a significant spatial or temporal change within the
video. More precisely, an \textit{I} frame is placed where one of the
following conditions is met:
\begin{itemize}
    \item The frame remarkably differs from the previous frame.
    \item One second has passed from the previous \textit{I} frame.
\end{itemize}

We explain the steps to obtain enhanced video frames using \method{} below.

\textbf{Step 1.} Apply NATLE to an \textit{I} frame:
\begin{equation}\label{eq:natle_I}
I_{enhanced}, \widehat{L}_{I}, L_{I}, \widehat{R}_{I}, R_{I} = \mli{NATLE(I)},
\end{equation}
where $I$ and $I_{enhanced}$ are the low-light and enhanced
keyframes, respectively. The rest of the parameters, i.e.,
$\widehat{L}_{I}$, $L_{I}$, $\widehat{R}_{I}$ and $R_{I}$, are the
results of intermediate steps in NATLE as described in Section
\ref{subsec:NATLE}.

\textbf{Step 2.} Learn two ridge regressions mapping
$\widehat{L}_{I}$ and $\widehat{R}_{I}$ to $L_{I}$ and $R_{I}$,
respectively. To be more precise, we look for $W_{L}$ and $W_{R}$ 
to solve the following optimization problems:
\begin{equation}\label{eq:ridge_regression_L}
\min_{W_{l}} ||l_{I} - \widehat{l}_{I} W_{l}||^2_2 + \alpha 
||W_{l}||^2_2,
\end{equation}
\begin{equation}\label{eq:ridge_regression_R}
\min_{W_{r}} ||r_{I} - \widehat{r}_{I} W_{r}||^2_2 + \alpha 
||W_{r}||^2_2,
\end{equation}
where $l_{I} \in \mathbb{R}^{n\times 1}$ is the vectorized
form of $L_{I}$ with $n$ pixels. $\widehat{l}_{I} \in
\mathbb{R}^{n\times 25}$ denotes $5\times 5$ neighborhoods of each pixel
in $\widehat{L}_{I}$. The solution, $W_{l} \in \mathbb{R}^{25\times 1}$,
maps each $5\times 5$ patch in $\widehat{L}_{I}$ to the corresponding
center pixel in $L_I$. The same process and notation is used for
Eq. (\ref{eq:ridge_regression_R}).

\textbf{Step 3.} Compute $\widehat{L}_{P}$ of \textit{B/P}
frames using Eq.~(\ref{eq:L_hat}). Obtaining $\widehat{L}$ by NATLE is
computationally inexpensive. Hence, we keep it the same while enhancing
\textit{B}/\textit{P} frames.

\textbf{Step 4.} Compute $L_{P}$ using the ridge regressor
$W_l$ learned in Step 2:
\begin{equation}\label{eq:ridge_regressor_L}
l_{P} = \widehat{l}_{P} W_l,
\end{equation}
where $\widehat{l}_{P} \in \mathbb{R}^{n\times 25}$ denotes
$5\times 5$ neighborhoods of each pixel in $\widehat{L}_{P}$. $l_{P} \in
\mathbb{R}^{n\times 1}$ is the vectorized form of $L_{P}$ with $n$
pixels. We reshape $l_{P}$ vector to obtain $L_{P}$ matrix.

\textbf{Step 5.} Compute $\widehat{R}_{P}$ for \textit{B/P}
frames using Eq.~(\ref{eq:Rhat}).

\textbf{Step 6.} Compute $R_{P}$ using the ridge regressor
$W_r$ learned in Step 2; namely,
\begin{equation}\label{eq:ridge_regressor_R}
r_{P} = \widehat{r}_{P} W_r,
\end{equation}
where $\widehat{r}_{P} \in \mathbb{R}^{n\times 25}$ denotes
$5\times 5$ neighborhoods of each pixel in $\widehat{R}_{P}$. $r_{P} \in
\mathbb{R}^{n\times 1}$ is the vectorized form of $R_{P}$ with $n$
pixels. We reshape $r_{P}$ to obtain $R_{P}$.

\textbf{Step 7.} Apply gamma correction to $L_{P}$ for 
illumination adjustment. The final enhanced \textit{B/P} frame 
is computed using Eq.~(\ref{eq:gamma}).

We perform Steps 1 and 2 on the \textit{I} frames and Steps 3
to 7 on the subsequent \textit{B/P} frames. Once a new \textit{I} frame
is encountered, we repeat Steps 1 and 2 and continue. This setting
ensures that the self-supervised learning from \natle{} is being updated
frequently enough to keep up with any significant temporal changes.

\begin{figure*}[t]
     \centering
     \begin{subfigure}[b]{0.23\textwidth}
         \centering
         \includegraphics[width=\textwidth]{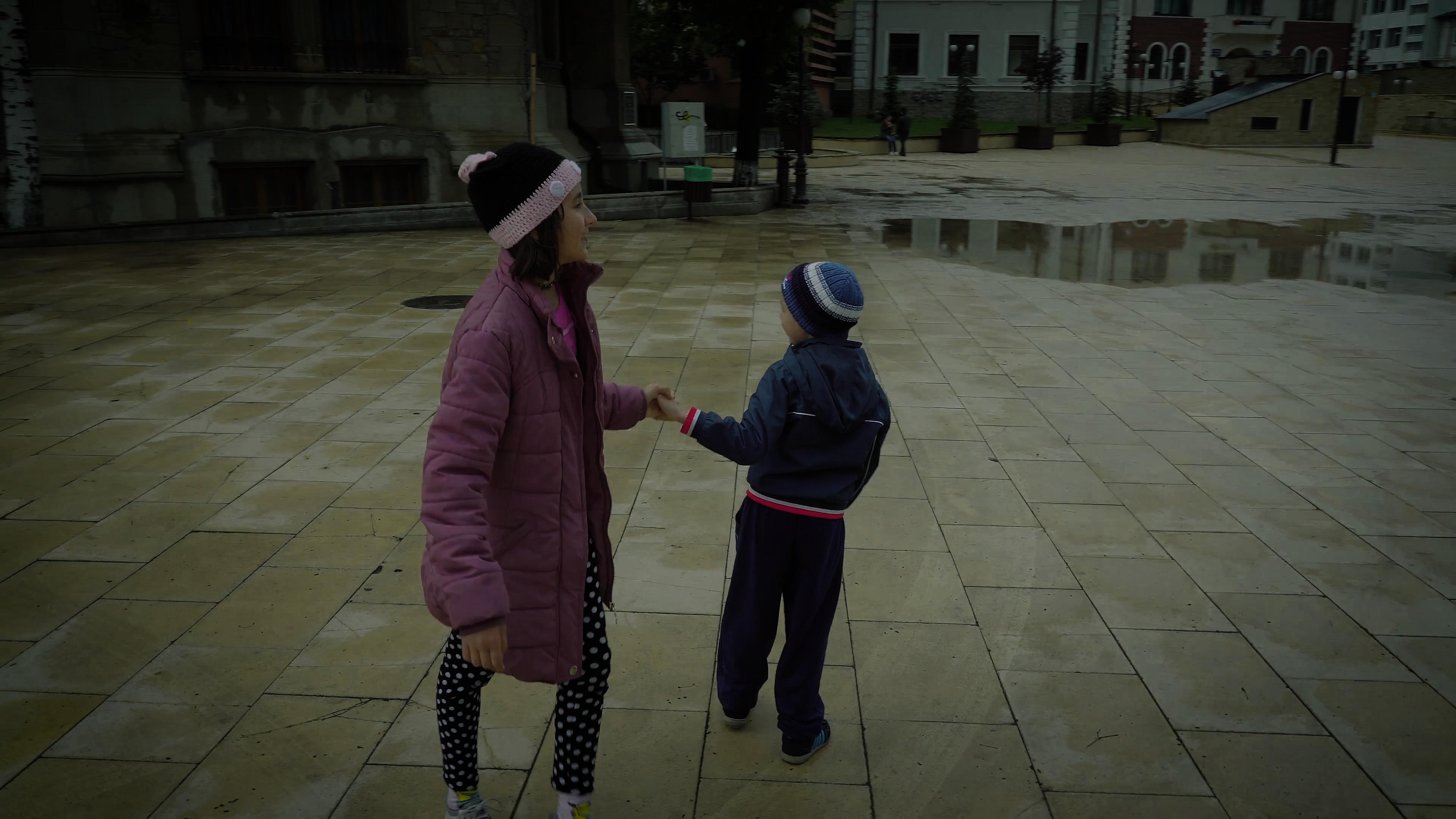}
         \caption{clean dark input}
     \end{subfigure}
     \begin{subfigure}[b]{0.23\textwidth}
         \centering
         \includegraphics[width=\textwidth]{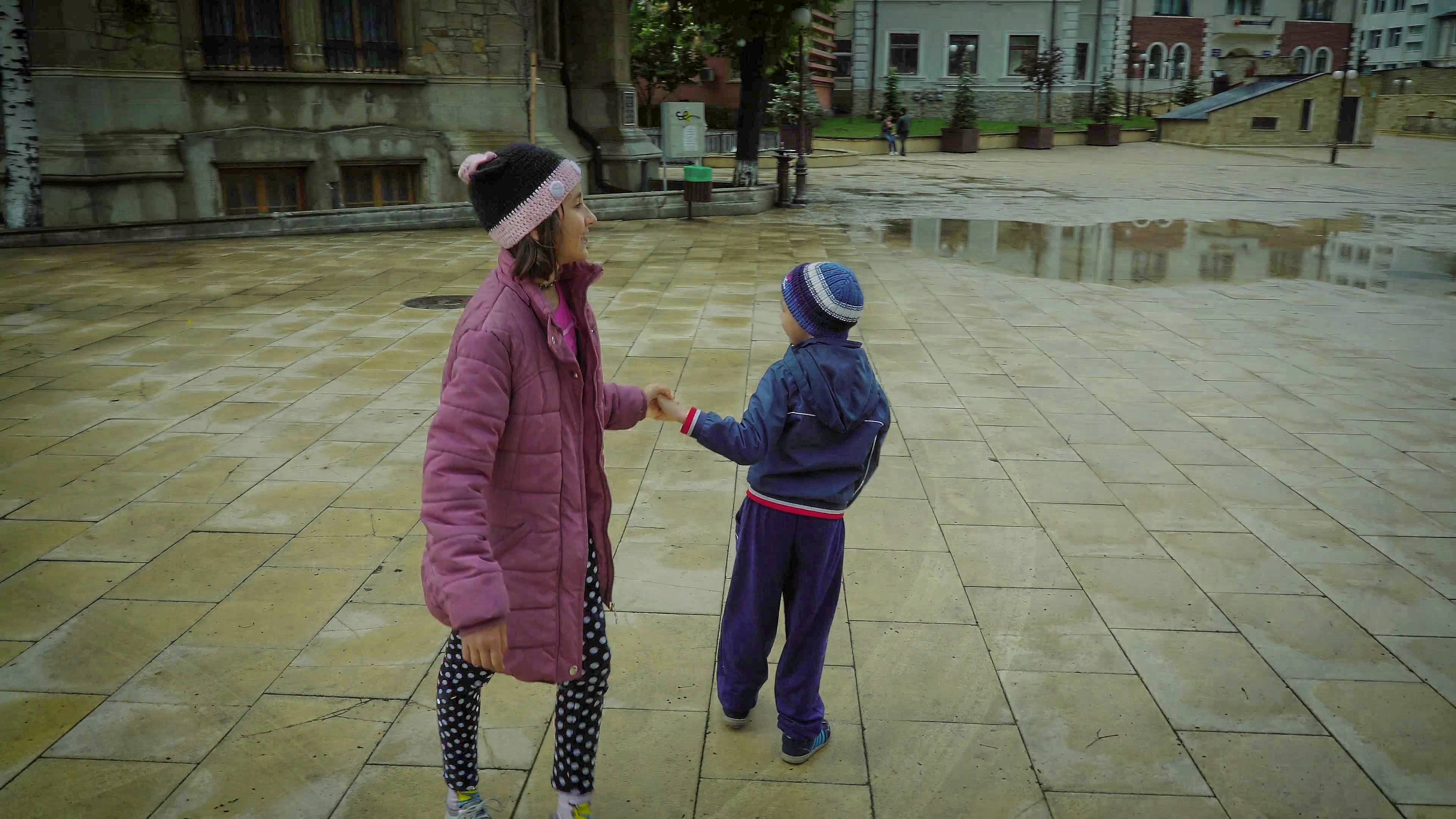}
         \caption{LIME \cite{guo2016lime}}
     \end{subfigure}
     \begin{subfigure}[b]{0.23\textwidth}
         \centering
         \includegraphics[width=\textwidth]{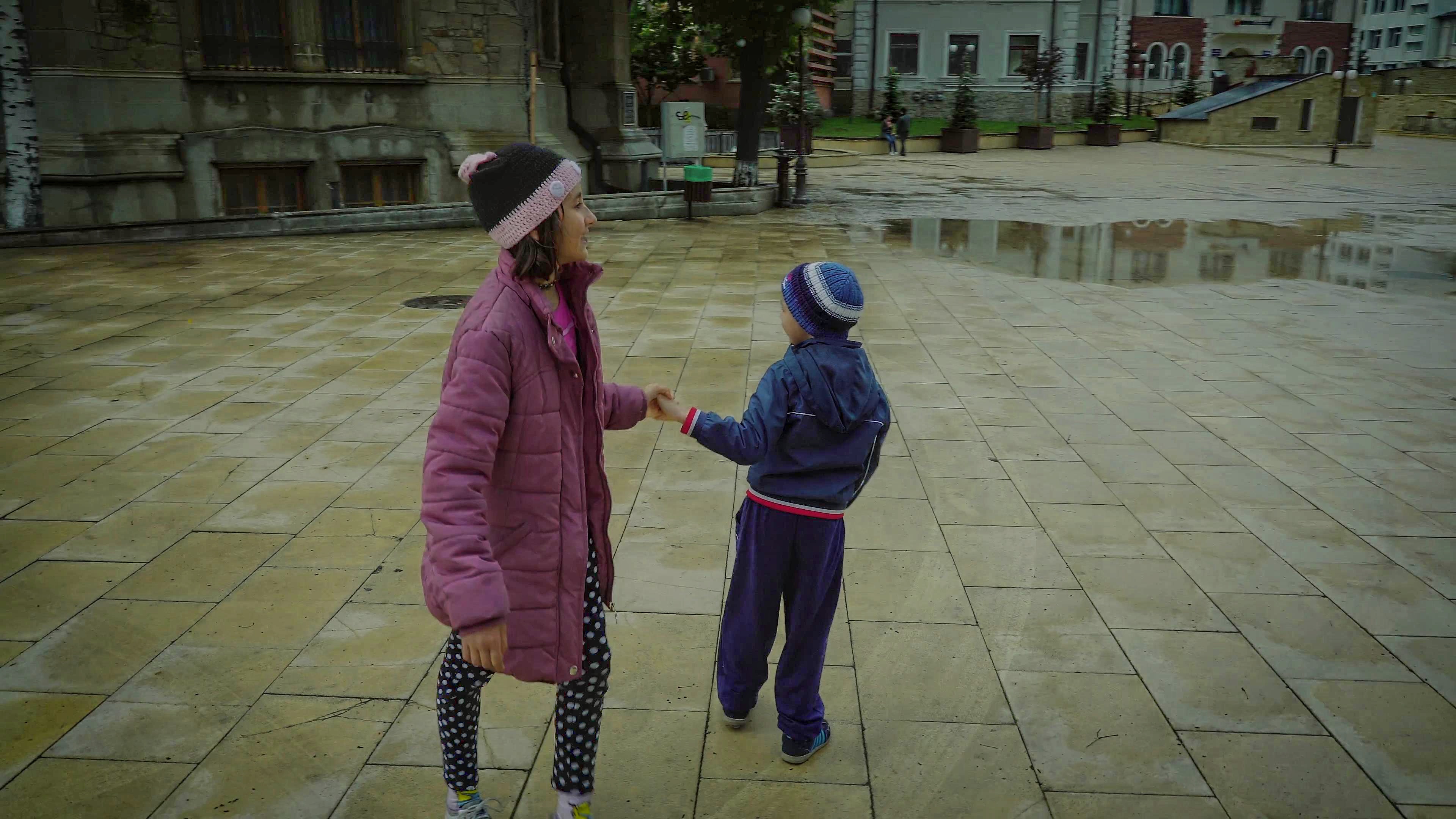}
         \caption{Dual \cite{zhang2019dual}}
     \end{subfigure}
     \begin{subfigure}[b]{0.23\textwidth}
         \centering
         \includegraphics[width=\textwidth]{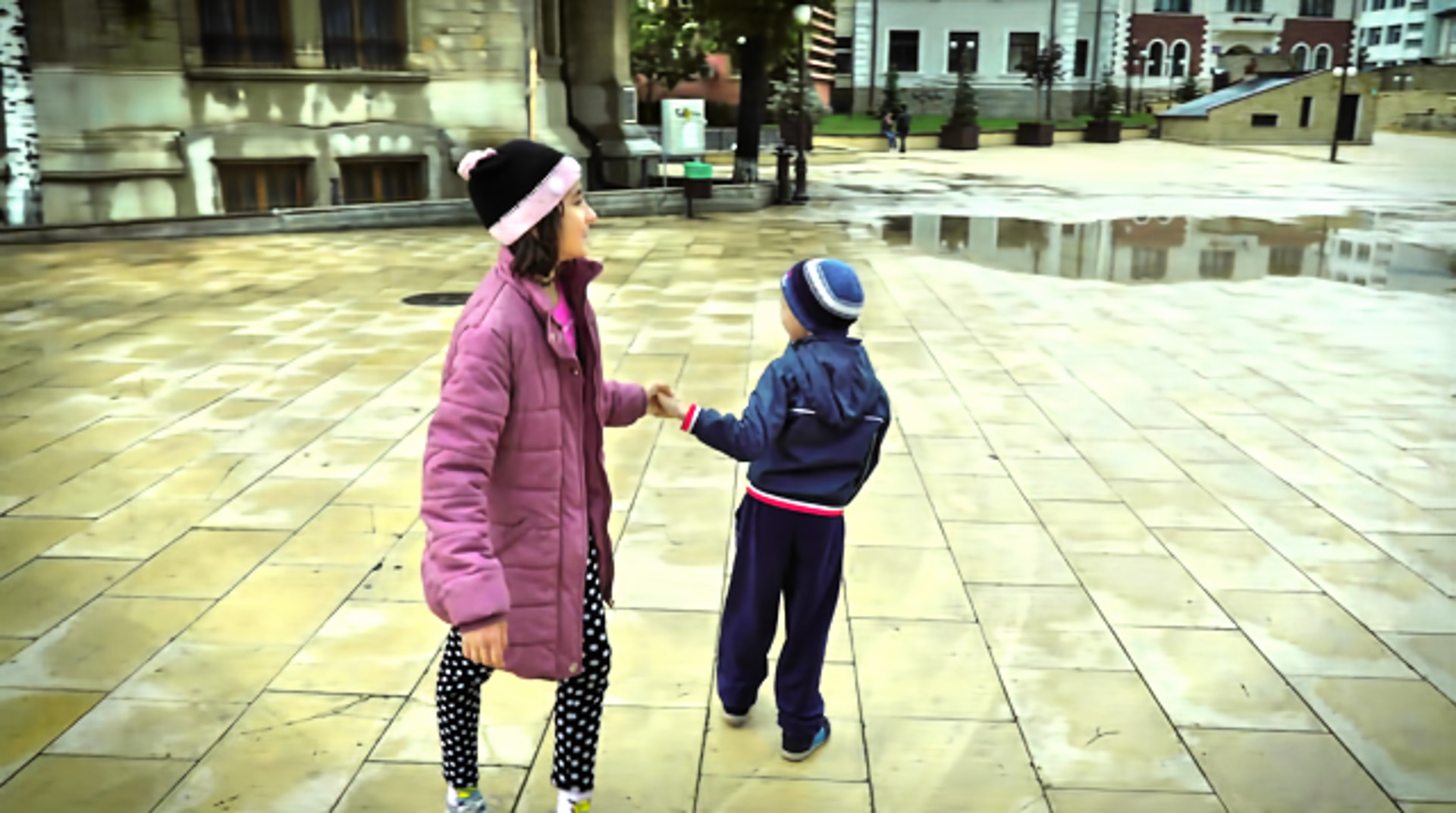}
         \caption{DRP \cite{liang2022self}}
     \end{subfigure}
     \begin{subfigure}[b]{0.23\textwidth}
         \centering
         \includegraphics[width=\textwidth]{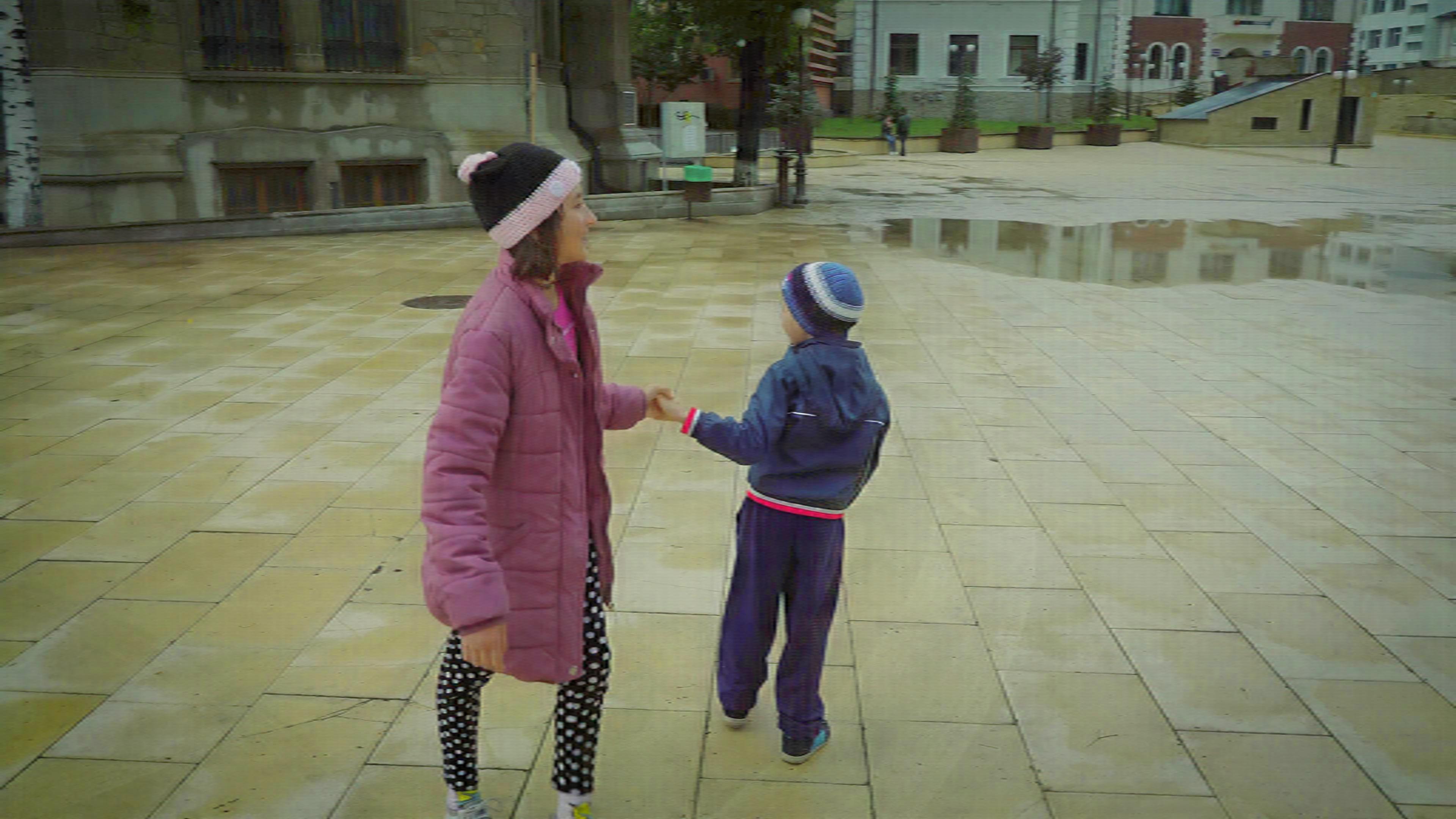}
         \caption{SDSD \cite{wang2021seeing}}
     \end{subfigure}
          \begin{subfigure}[b]{0.23\textwidth}
         \centering
         \includegraphics[width=\textwidth]{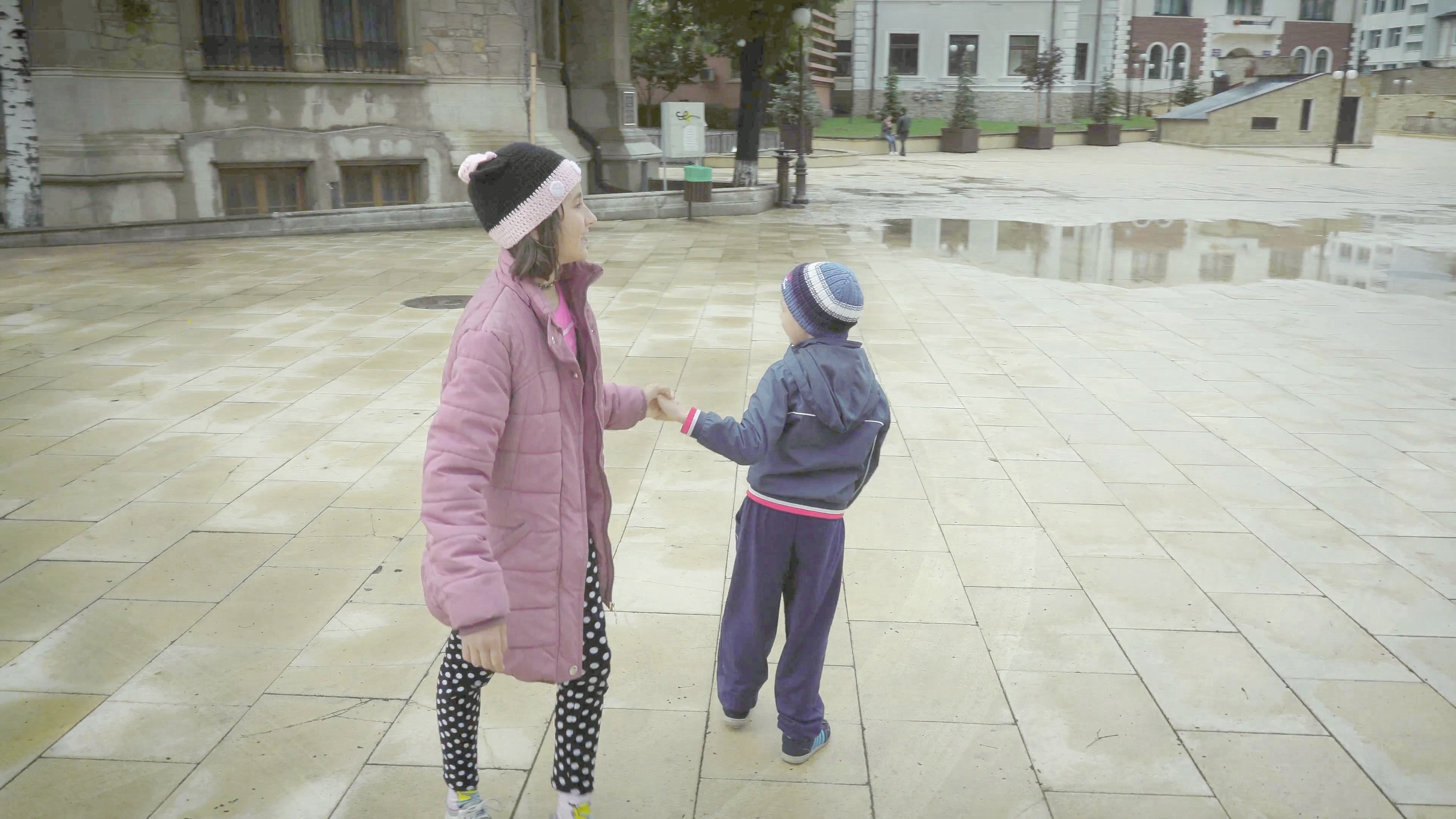}
         \caption{StableLLVE \cite{zhang2021learning}}
     \end{subfigure}
    \begin{subfigure}[b]{0.23\textwidth}
         \centering
         \includegraphics[width=\textwidth]{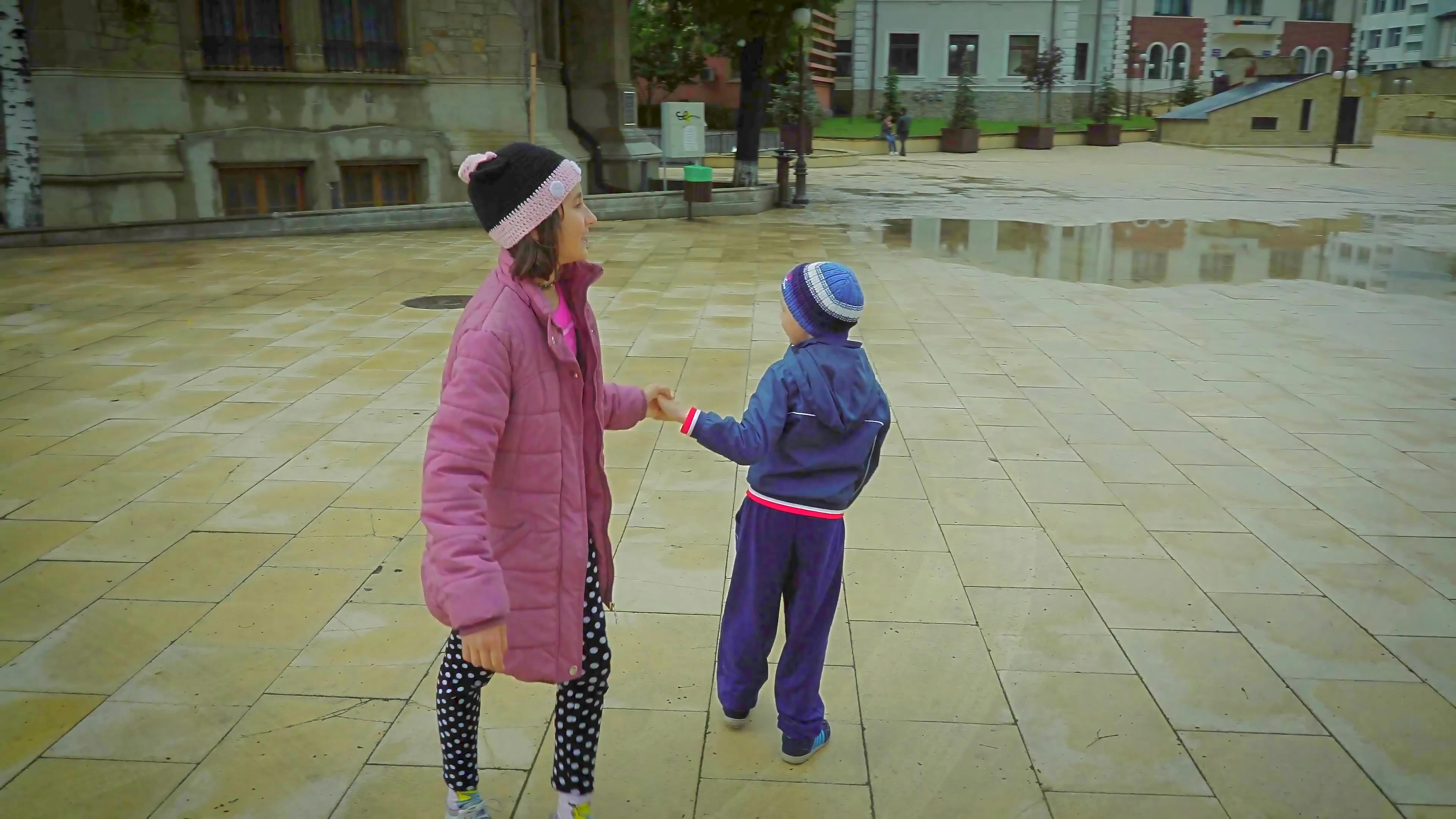}
         \caption{Ours}
     \end{subfigure}
      \begin{subfigure}[b]{0.23\textwidth}
         \centering
         \includegraphics[width=\textwidth]{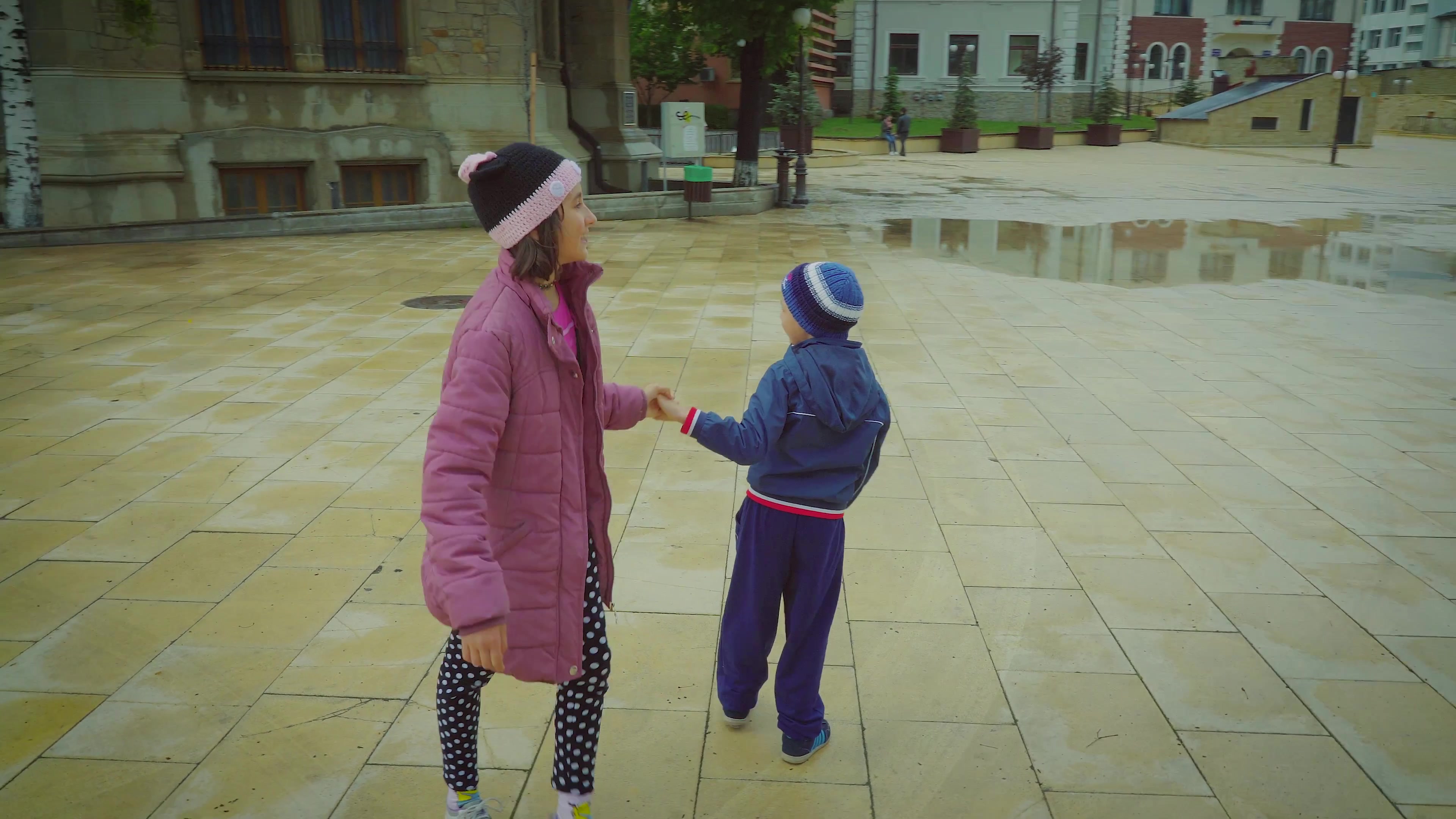}
         \caption{Ground-truth}
     \end{subfigure}
     
    \caption{Visual comparison between our method and prior work on clean-darkened video frames from DAVIS dataset.  }
    \label{fig:visual_clean}
\end{figure*}


\begin{table}[t]
\caption{Quantitative comparison for enhancing clean dark videos. The best scores are indicated in \textbf{bold}.}\label{tab:clean}
\centering
\resizebox{0.7\columnwidth}{!}{
\begin{tabular}{l|cccc}
Method & PSNR$\uparrow$ & SSIM$\uparrow$ & AB(Var)$\downarrow$ & MABD$\downarrow$ \\ \hline
LIME \cite{guo2016lime} & 17.36 & 0.7386 & 9.65 & 0.37 \\
Dual \cite{zhang2019dual} & 18.12 & 0.8283 & 2.13 & 0.07 \\
MBLLEN \cite{lv2018mbllen} & 18.41  & 0.8100 & 77.24 & 1.95 \\
RetinexNet \cite{wei2018deep} & 19.78 & 0.8353 & 1.32 & 0.09\\
SID \cite{chen2018learning} & 22.95  & 0.9428 & 4.93 & 0.43 \\ 
DRP \cite{liang2022self} & 6.89  & 0.3160 & 6.73 & 0.52\\ \hline
MBLLVEN \cite{lv2018mbllen} & 24.50  & 0.9482 & 1.79 & 0.80 \\
SMOID \cite{jiang2019learning} & 24.85  & 0.9472 & \textbf{1.30} & 0.17\\ \hline
SFR \cite{eilertsen2019single} & 23.81 & 0.9413 & 2.14 & 0.11\\ 
BLIND \cite{lai2018learning} & 22.87 & 0.9344 & 8.66 & 0.43\\ 
StableLLVE \cite{zhang2021learning} & 24.07 & \textbf{0.9483} & 1.96 & 0.05\\ 
SDSD \cite{wang2021seeing} & 25.09 & 0.8783 & \textbf{0.98} & 0.01 \\ \hline
\method{} (Ours) & \textbf{28.85} & 0.9225 & 1.47 & \textbf{0.006 }\\ 
\end{tabular}
}
\end{table}

\begin{figure*}[t]
     \centering
     \begin{subfigure}[b]{0.23\textwidth}
         \centering
         \includegraphics[width=\textwidth]{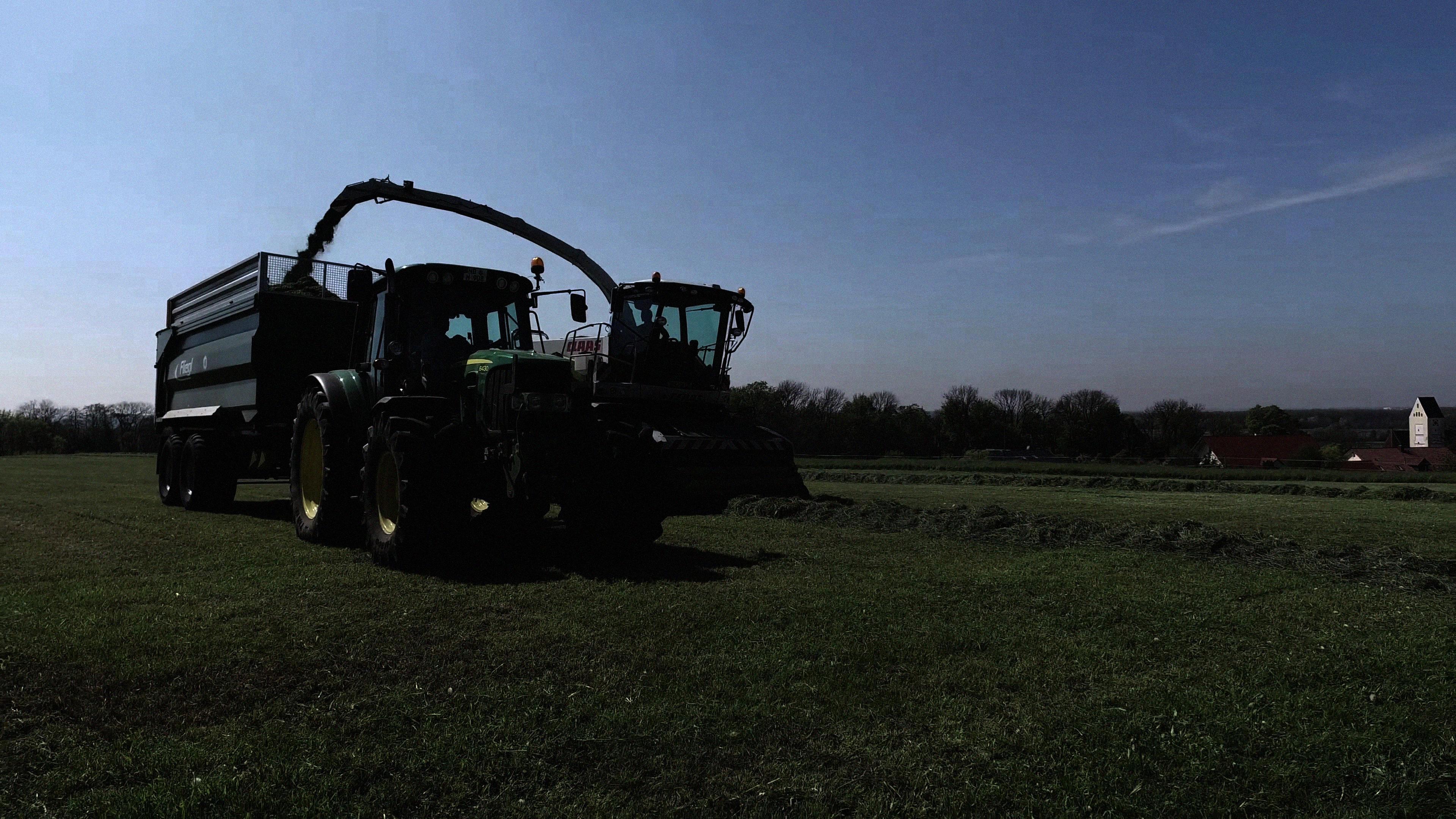}
         \caption{noisy dark input}
     \end{subfigure}
     \begin{subfigure}[b]{0.23\textwidth}
         \centering
         \includegraphics[width=\textwidth]{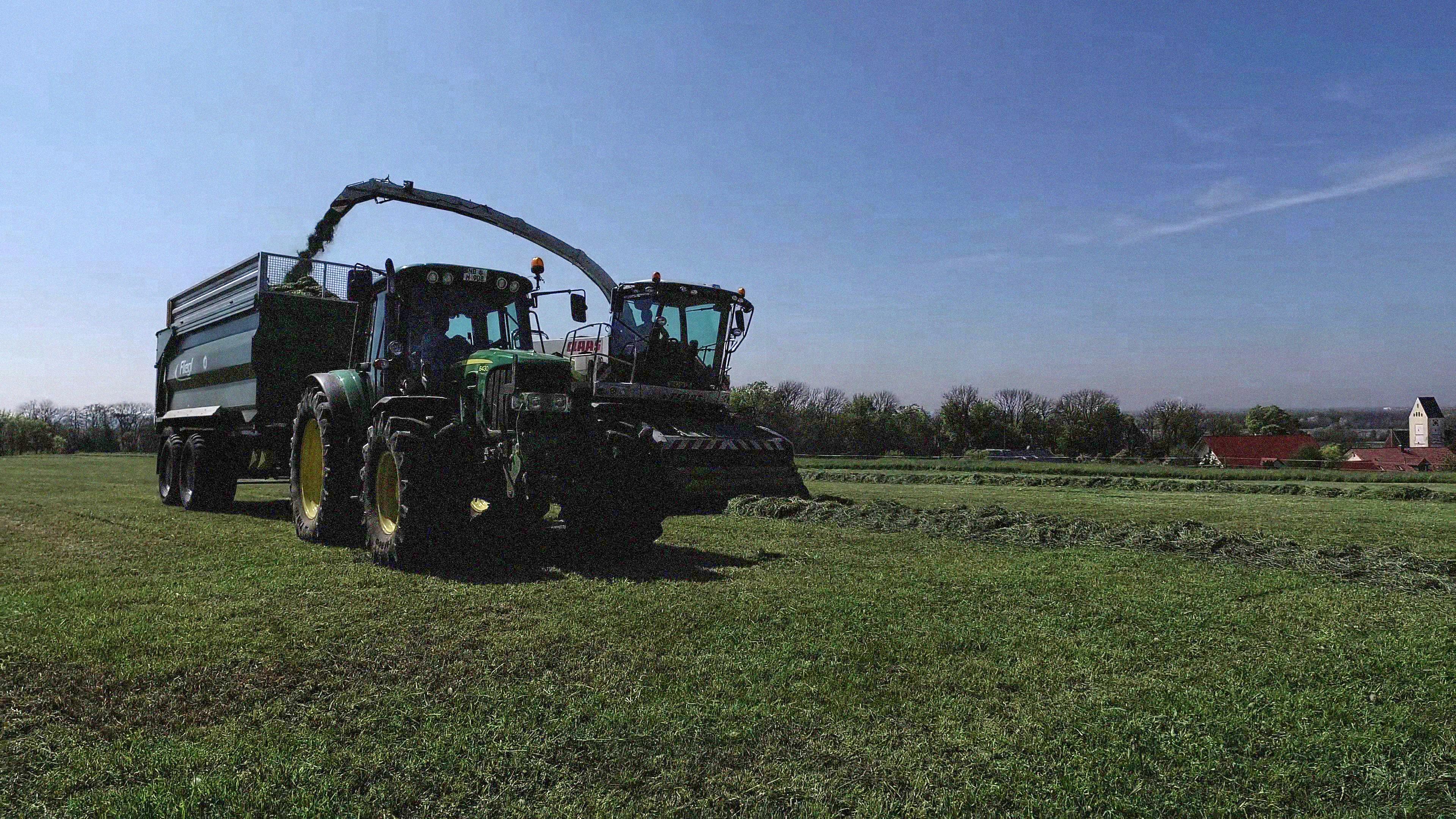}
         \caption{LIME \cite{guo2016lime}}
     \end{subfigure}
     \begin{subfigure}[b]{0.23\textwidth}
         \centering
         \includegraphics[width=\textwidth]{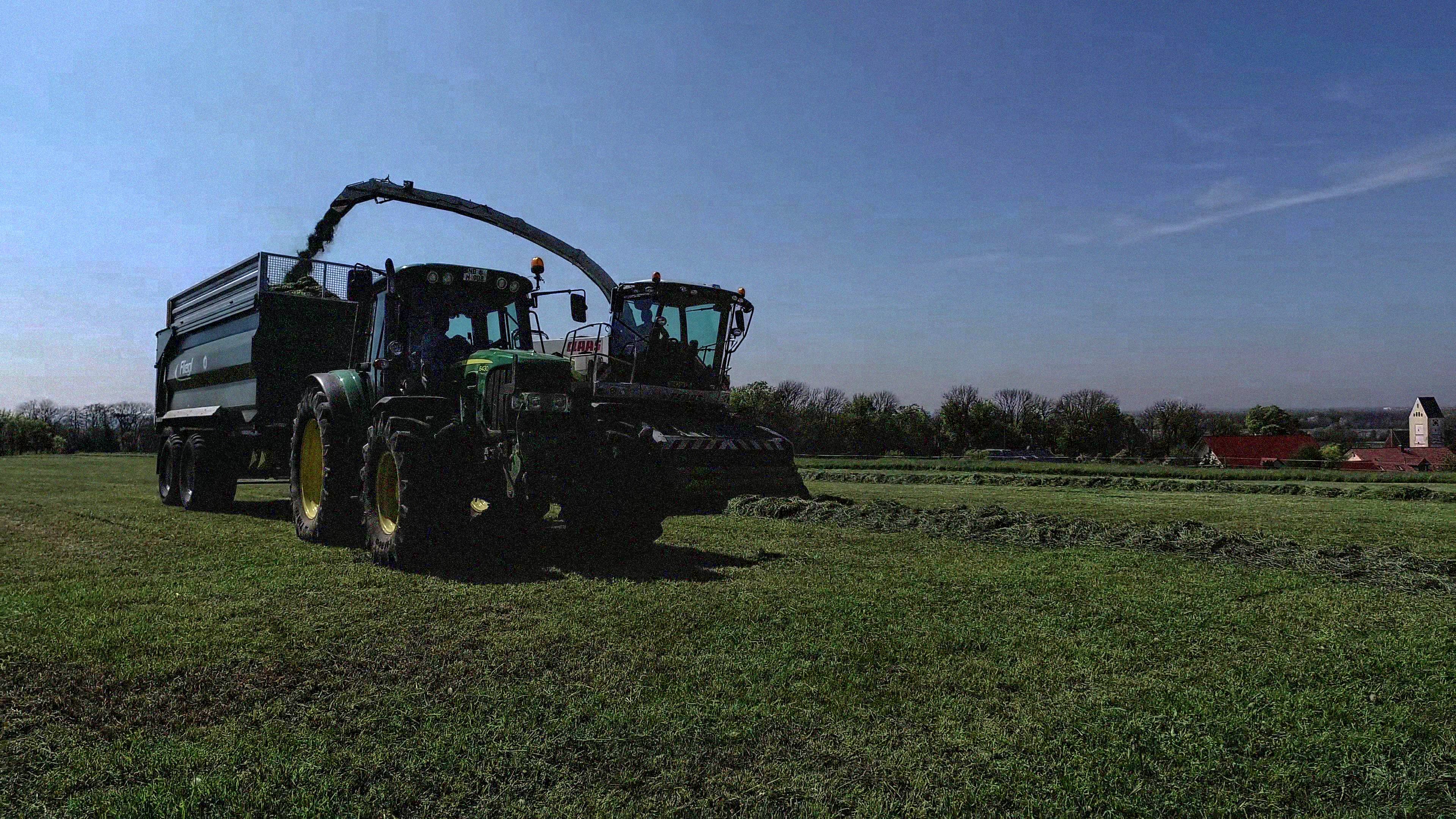}
         \caption{Dual \cite{zhang2019dual}}
     \end{subfigure}
     \begin{subfigure}[b]{0.23\textwidth}
         \centering
         \includegraphics[width=\textwidth]{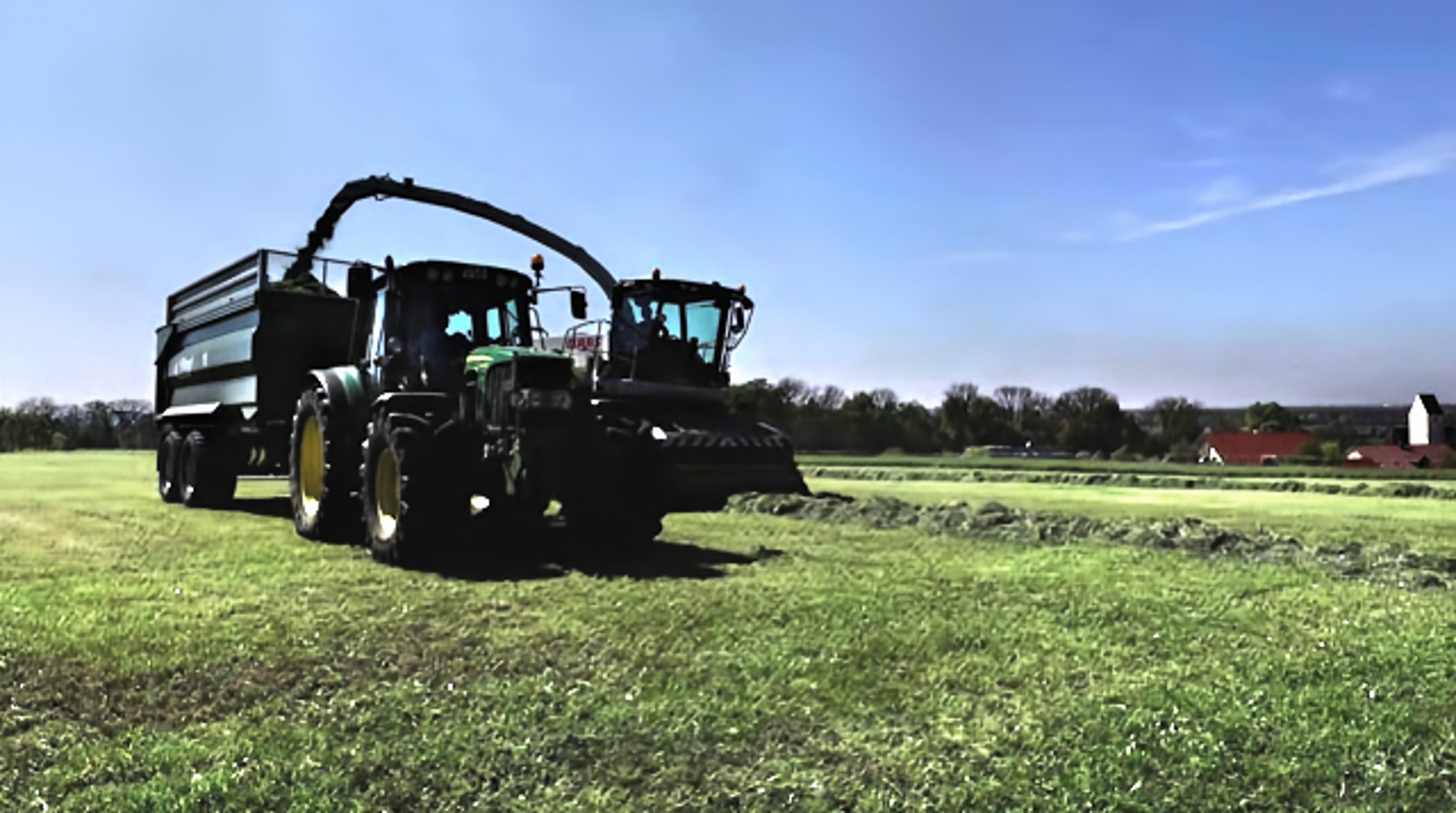}
         \caption{DRP \cite{liang2022self}}
     \end{subfigure}
     \begin{subfigure}[b]{0.23\textwidth}
         \centering
         \includegraphics[width=\textwidth]{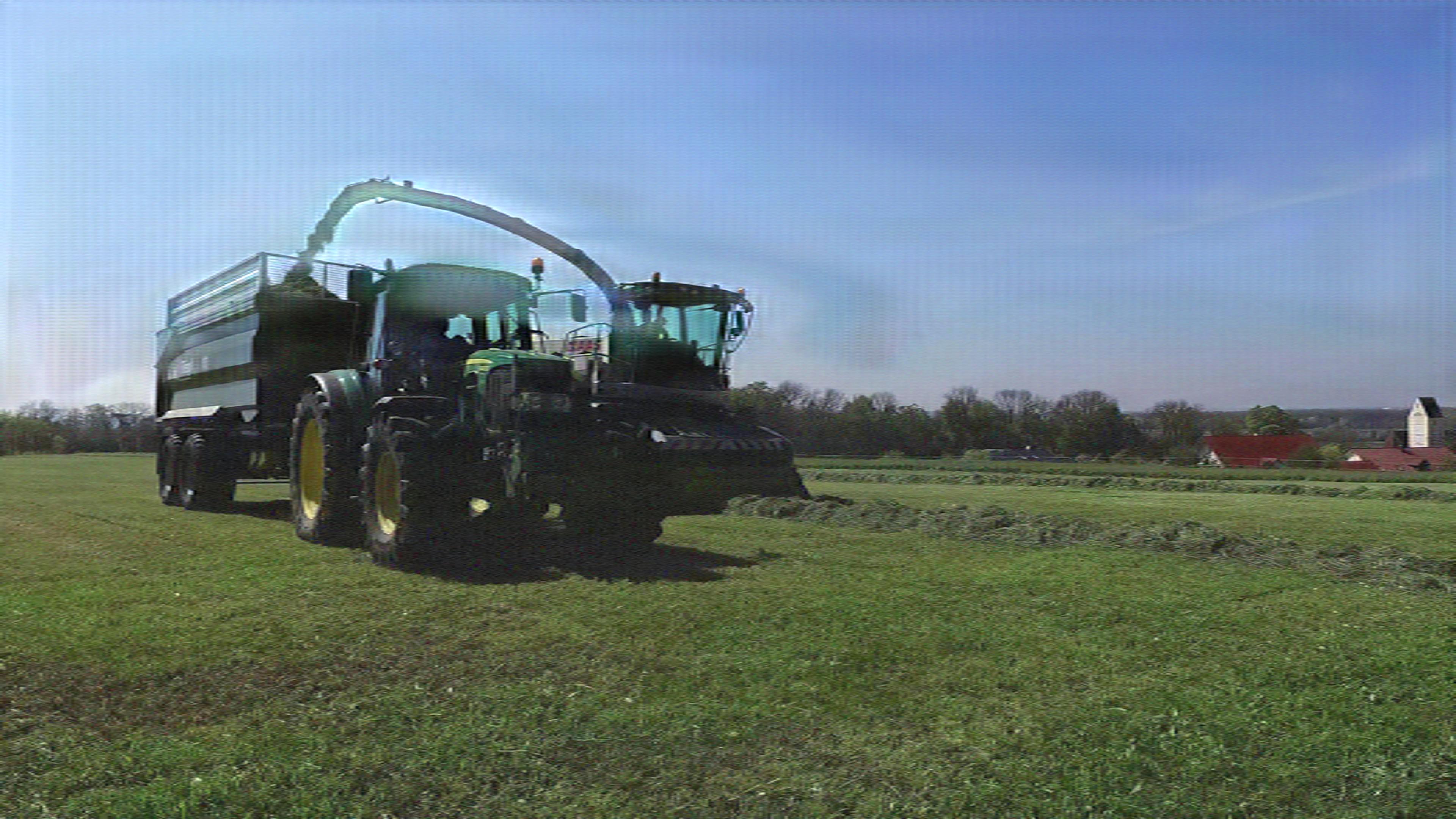}
         \caption{SDSD \cite{wang2021seeing}}
     \end{subfigure}
          \begin{subfigure}[b]{0.23\textwidth}
         \centering
         \includegraphics[width=\textwidth]{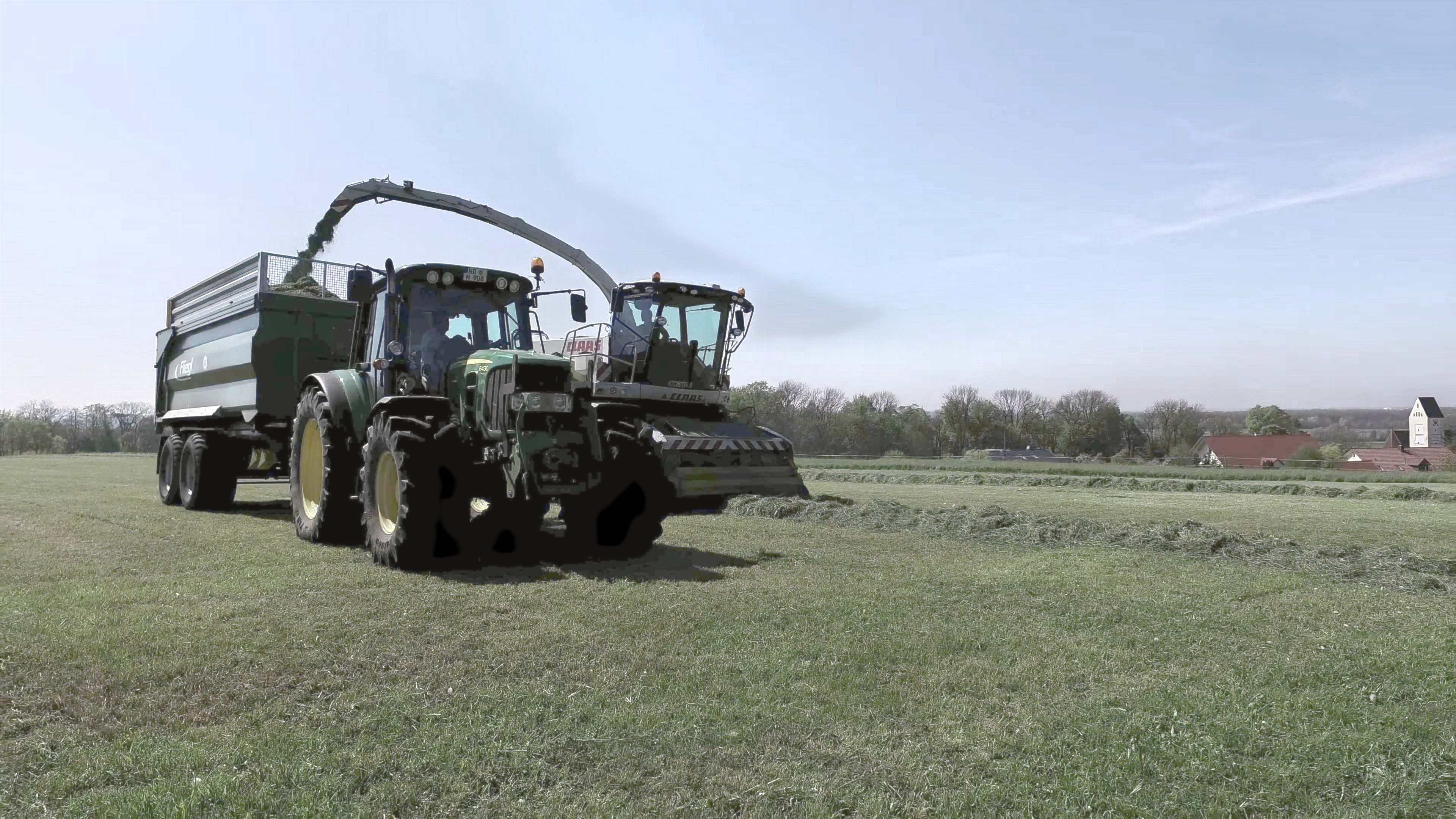}
         \caption{StableLLVE \cite{zhang2021learning}}
     \end{subfigure}
    \begin{subfigure}[b]{0.23\textwidth}
         \centering
         \includegraphics[width=\textwidth]{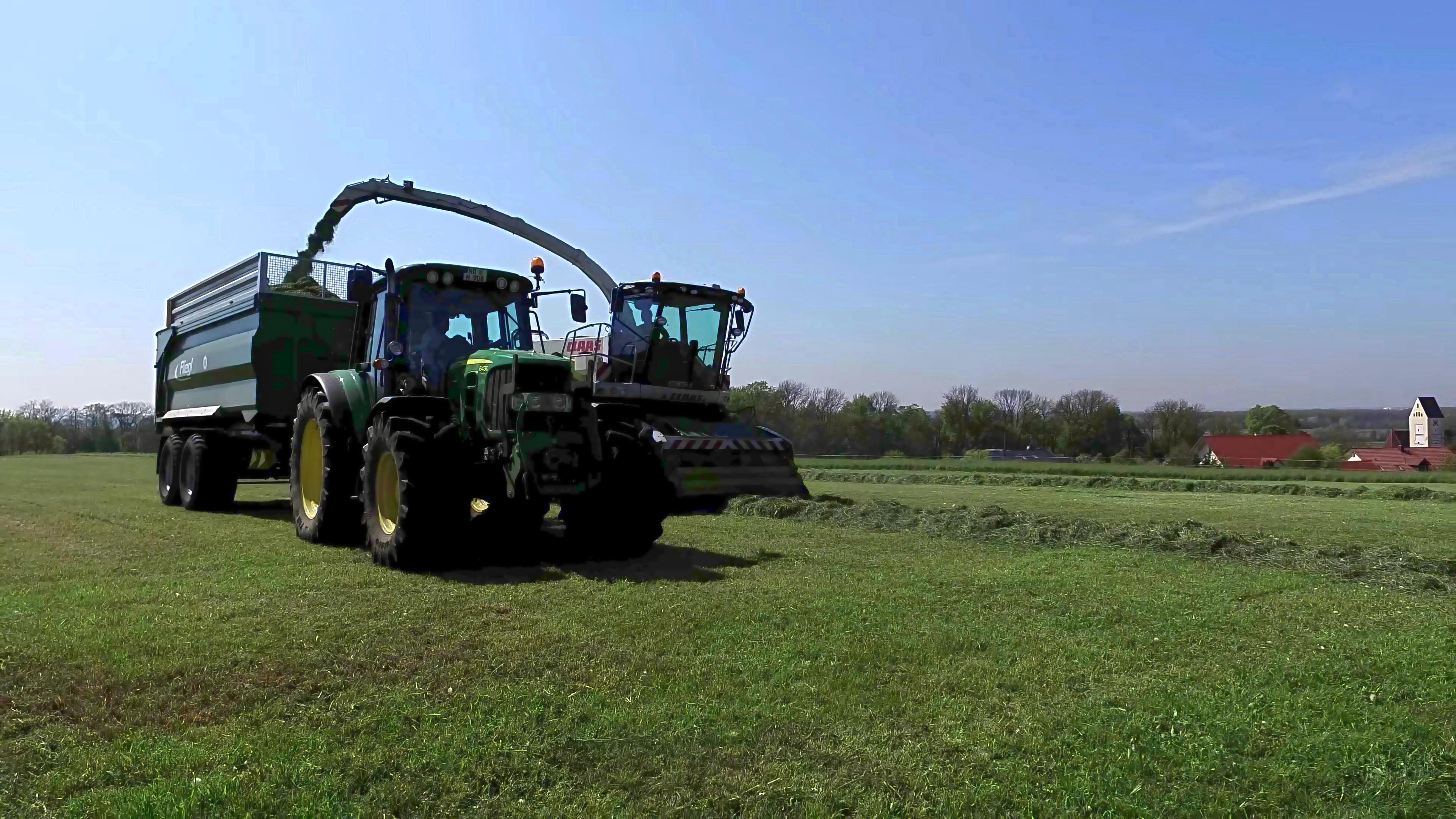}
         \caption{Ours}
     \end{subfigure}
      \begin{subfigure}[b]{0.23\textwidth}
         \centering
         \includegraphics[width=\textwidth]{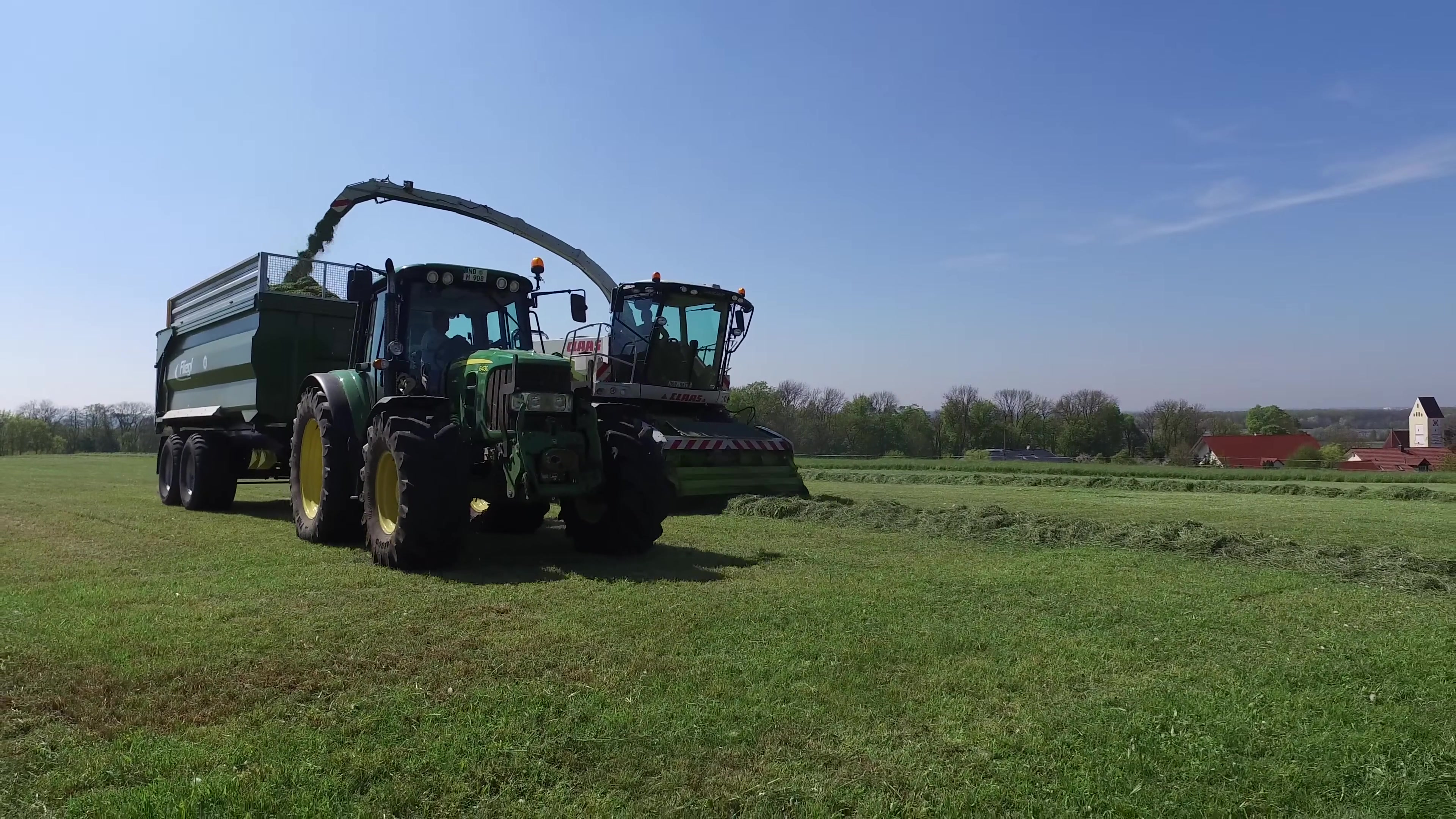}
         \caption{Ground-truth}
     \end{subfigure}
     
    \caption{Visual comparison between our method and prior work on noisy-darkened video frames from DAVIS dataset.  }
    \label{fig:visual_noisy}
\end{figure*}

\begin{table}[t]
\caption{Quantitative comparison for enhancing noisy dark videos. The best scores are indicated in \textbf{bold}.}\label{tab:noisy}
\centering
\resizebox{0.7\columnwidth}{!}{
\begin{tabular}{l|cccc}
Method & PSNR$\uparrow$ & SSIM$\uparrow$ & AB(Var)$\downarrow$ & MABD$\downarrow$ \\ \hline
LIME \cite{guo2016lime} & 16.43 & 0.4567 & 8.29 & 0.33 \\
Dual \cite{zhang2019dual} & 18.38 & 0.6073 & 2.14 & 0.22 \\
MBLLEN \cite{lv2018mbllen} & 18.38  & 0.7982 & 78.76 & 1.93 \\
RetinexNet \cite{wei2018deep} & 19.56 & 0.7475 & 1.45 & 0.09\\
SID \cite{chen2018learning} & 22.93  & 0.9253 & 4.03 & 0.39 \\ 
DRP \cite{liang2022self} & 5.55  & 0.4107 & 15.37 & 0.34\\ \hline
MBLLVEN \cite{lv2018mbllen} & 23.08  & 0.8839 & 2.81 & 1.02 \\
SMOID \cite{jiang2019learning} & 23.42  & 0.9212 & \textbf{0.82} & 0.17\\ \hline
SFR \cite{eilertsen2019single} & 22.82 & 0.9299 & 2.29 & 0.12\\ 
BLIND \cite{lai2018learning} & 22.94 & 0.9174 & 7.86 & 0.33\\ 
StableLLVE \cite{zhang2021learning} & 24.01 & \textbf{0.9305} & 3.00 & 0.10\\ 
SDSD \cite{wang2021seeing} & 22.27 & 0.8051 & 1.35 & 0.03 \\ \hline
\method{} (Ours) & \textbf{27.06} & 0.8202 & 1.18 & \textbf{0.01} \\ 
\end{tabular}
}
\end{table}

\begin{figure*}
    \centering
    \includegraphics[width=\textwidth]{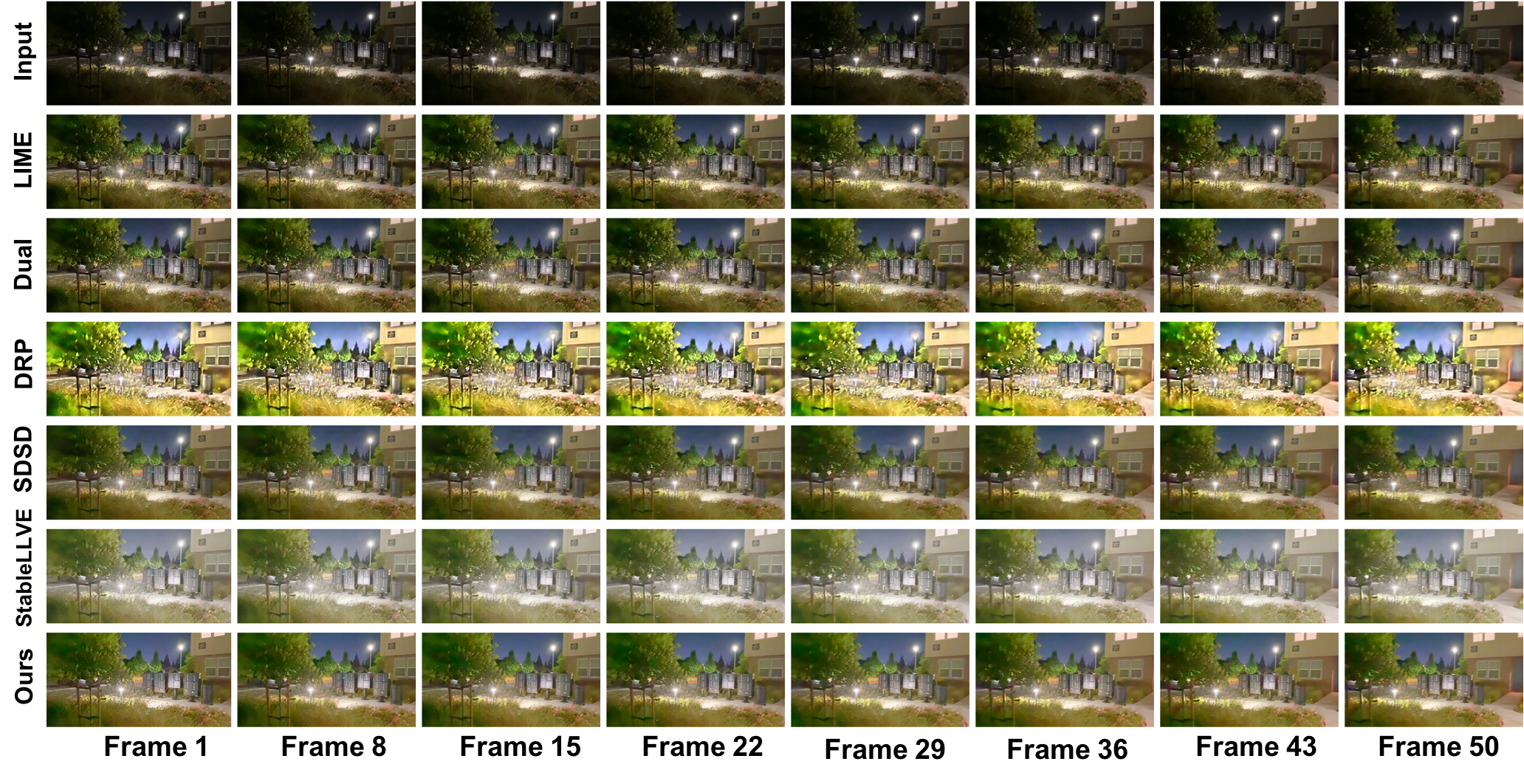}
    \caption{Visualization of video frames from a real-world dark video from LoLi-Phone dataset.}
    \label{fig:visual_real}
\end{figure*}

\section{Experiments}\label{sec:experiments}

\subsection{Experimental Setup}\label{subsec:setup}

We conduct extensive qualitative and quantitative experiments to evaluate our method and show its effectiveness. In our experiments, we use the DAVIS video dataset \cite{pont20172017, caelles20192019} as the ground truth. DAVIS offers two resolutions, 480P and full resolution. We use all full resolution videos from 2017 and 2019 challenges. Following \cite{lv2021attention}, we synthesize dark videos by darkening the normal-light frames of DAVIS dataset with gamma correction and linear scaling:
\begin{equation}\label{eq:darken}
    x = \mathcal{B}\times(\mathcal{A}\times y)^\gamma
\end{equation}
where $y$ is the ground-truth (normal light) frame, $x$ is the darkened frame, $\mathcal{A}$ and $\mathcal{B}$ are linear scaling factors and sampled from uniform distributions $U(0.9,1)$ and $U(0.5,1)$, respectively, and $\gamma$ is the gamma correction factor which is sampled from $U(2,3.5)$. 

We also synthesized the noisy version of the dark frames. Following \cite{zhang2021learning}, we add both Poisson and Gaussian noise to the low-light frames:
\begin{equation}\label{eq:noise}
    n = \mathcal{P}(\sigma_p)+\mathcal{N}(\sigma_g)
\end{equation}
where $\sigma_p$ and $\sigma_g$ are parameters of Poisson noise and Gaussian noise, respectively. They are both sampled from a uniform distribution $U(0.01, 0.04)$. We acquire two sets of videos, namely darkened and noisy-darkened videos. Our goal is to enhance these videos and assess them qualitatively and quantitatively.

\subsection{Visual Comparison}\label{subsec:visual}
We first provide visual analysis on an exemplary video frame from the
clean-dark and noisy-dark datasets in Figures ~\ref{fig:visual_clean}
and \ref{fig:visual_noisy}, respectively. We observe that methods not
based on deep learning (LIME and DUAL) do not add artifacts to the
frame, but the resulting enhanced frame still lacks lightness.
Among prior methods based on deep learning, DRP
\cite{liang2022self} is a self-supervised method that gives nice
enhancement results. While DRP adds colorful textures to the enhanced
images, the results tend to be slighly different with the ground truth
and have artifacts in some regions. SDSD \cite{wang2021seeing} is a
supervised method which is fine-tuned to the DAVIS dataset. SDSD tends
to add artifacts to enhanced images. This issue is more noticable in
Figure~\ref{fig:visual_noisy}. StableLLVE~\cite{zhang2021learning} is a
supervised method trained on the DAVIS dataset. The enhancement results
by StableLLVE have a pale color. Our method achieves enhanced frames
that are fairly close to the ground-truth and avoids adding artifacts or
changing the coloring of the image. 

The exemplary input dark frames in Figures \ref{fig:visual_clean} and
\ref{fig:visual_noisy} were created synthetically. Next, we examine our
framework on a real-world video randomly selected from the LoLi-Phone
dataset~\cite{li2021low}. Note that there is no ground-truth video in
this case. We present the enhanced frames corresponding to our work and
related work in Figure~\ref{fig:visual_real}. We have a similar
observation to that of the synthetic dataset. Our method is capable of
achieving an image with a natural lightness while keeping the coloring
and visual content intact.

\subsection{Quantitative Evaluation}\label{subsec:scores}

We use four quantitative metrics to evaluate the performance of our
method and compare it with prior work. First two metrics are
Peak-Signal-to-Noise Ratio (PSNR) and Structural-SIMilarity (SSIM)
\cite{wang2004image}, which we apply on all frames of the videos. We
also use AB(Var) from \cite{lv2018mbllen} and Mean Absolute Brightness
Difference (MABD) from \cite{jiang2019learning} to assess temporal
stability and its consistency with those of the ground truth. Table
\ref{tab:clean} and Table \ref{tab:noisy} show comparisons between our
method and prior work for the clean and noisy datasets, respectively.
In Table \ref{tab:clean} and \ref{tab:noisy}, we take the
scores of all prior work except for Dual \cite{zhang2019dual}, DRP
\cite{liang2022self} and SDSD \cite{wang2021seeing} from
\cite{zhang2021learning}.  For DUAL (traditional method) and DRP
(self-supervised method), we use their public codes to enhance the
images. For SDSD, we fine-tune their pre-trained model on the DAVIS
dataset. We then use the fine-tuned SDSD model to enhance the images. We
then calculate the scores of these three methods and report them in
Tables~\ref{tab:clean} and \ref{tab:noisy}.


\begin{table}[t]
\caption{Average runtime (in seconds) comparison per RGB frame
of size $530\times 942$ pixels.}\label{tab:runtime}
\centering
\resizebox{0.38\columnwidth}{!}{
\begin{tabular}{l|cc}
Method & CPU & GPU \\ \hline
LIME \cite{guo2016lime} & 6.60 & N/A \\
Dual \cite{zhang2019dual} & 13.20 & N/A \\
DRP \cite{liang2022self} & 2760  & 2728 \\  
StableLLVE \cite{zhang2021learning} & 0.063 & 0.057 \\ 
SDSD \cite{wang2021seeing} & 0.307 & 0.261 \\
\method{} (Ours) & 0.980 & 0.322 \\ 
\end{tabular}
}
\end{table}

\begin{table}[t]
\caption{FLOPs comparison per pixel.}\label{tab:flop}
\centering
\resizebox{0.43\columnwidth}{!}{
\begin{tabular}{l|cc}
Method & FLOPs & Ratio\\ \hline
StableLLVE \cite{zhang2021learning} & 51.19 K & $7.20\times$ \\ 
SDSD \cite{wang2021seeing} & 233.45 K & $32.83\times$ \\
\method{} (Ours) & 7.11 K & $1\times$ \\ 
\end{tabular}
}
\end{table}


\subsection{Computational Complexity}\label{subsec:flop}

In this section, we calculate the runtime and FLOPs (FLoating
Point Operations) of \method{} and prior work to offer a comparison on
the computational complexity.

Table \ref{tab:runtime} shows the runtime comparison between
different methods on CPU and GPU resources. We measure the average
runtime of different methods for enhancing an RGB frame of size
$530\times 942$ on the CPU resource of \textit{Intel Xeon 6130} and the 
GPU resource of \textit{NVIDIA Tesla V100}. Since LIME and Dual only have
CPU implementations, their GPU runtimes are not mentioned in the table.
Table \ref{tab:runtime} shows that \method{}'s runtime is better than
LIME, Dual and DRP. Specifically, \method{} is more than 2500 times
faster than the self-supervised DRP method.

Table \ref{tab:flop} shows a comparison on the number of FLOPs
per pixel between StableLLVE, SDSD and \method{}. We use a FLOP counter
tool\footnote{\url{https://github.com/sovrasov/flops-counter.pytorch}}
\cite{ptflops} for PyTorch to calculate FLOPs of StableLLVE and SDSD. We
did not find a tool to measure the number of FLOPs of (non-PyTorch)
LIME, Dual, and DRP methods. For \method{}, we calculate the number of
the FLOPs manually \cite{cools2017communication}.

Table \ref{tab:flop} shows that \method{} has a significantly
lower number of FLOPs than StableLLVE and SDSD. Our explanation for the
lower runtime of these two methods in Table \ref{tab:runtime} is that
their implementations in PyTorch are very efficient. In contrast,
\method{} uses several libraries including SciPy in most of the
calculations. The latter is not as efficient as PyTorch. As a result,
\method{} has a slightly longer runtime despite its lower number of
FLOPs.

\subsection{User Study}\label{subsec:user}
\begin{figure}[t]
\centering
\includegraphics[width=0.8\columnwidth]{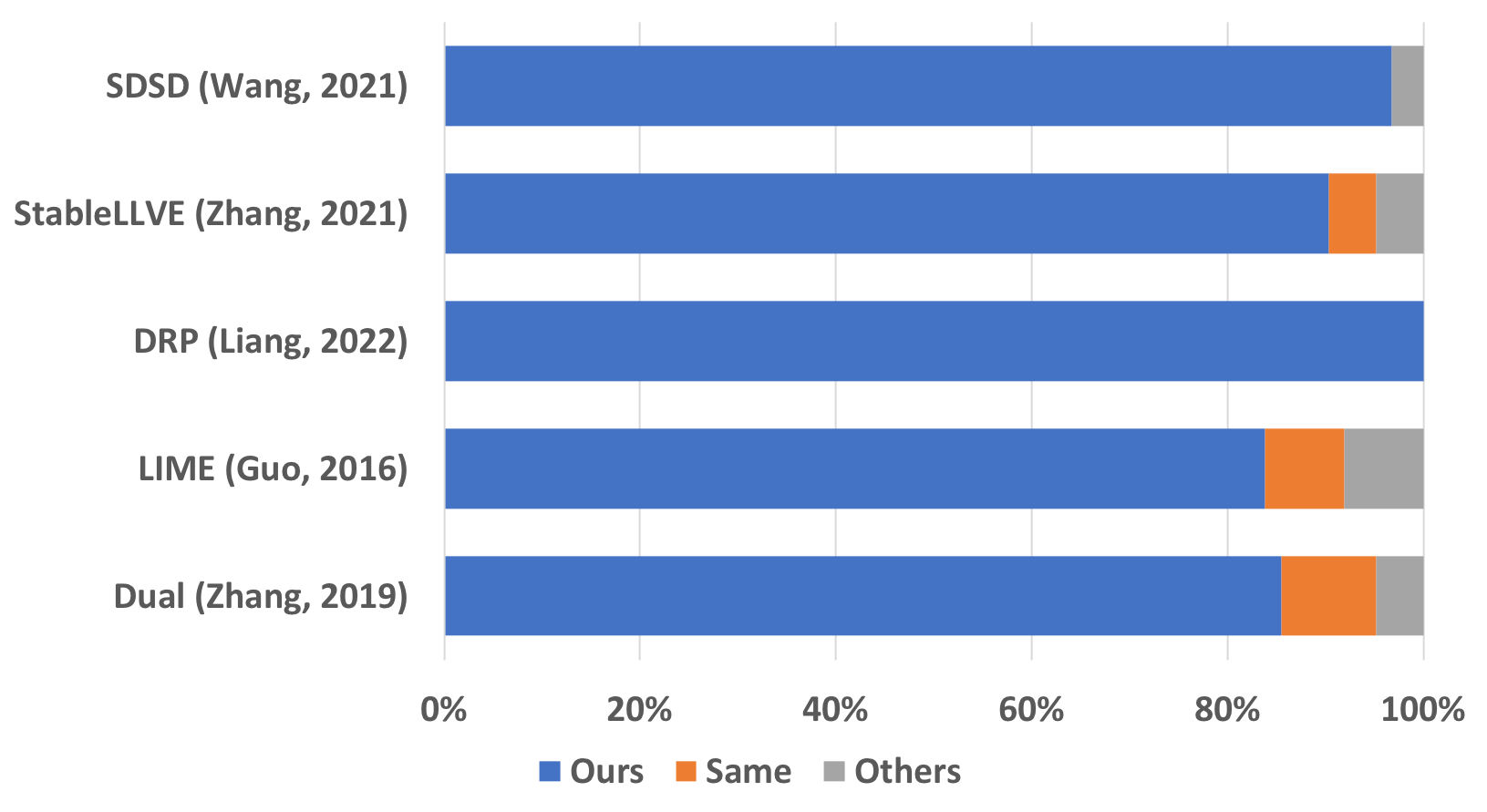}
\caption{User study} \label{fig:userstudy}
\end{figure}

To further demonstrate the effectiveness of our method, we conduct a user study with 31 participants. In this study, we have 10 blind A/B tests between our method and prior works. At each time, only 2 videos are shown to the user. The 10 videos are randomly selected for this study. Each of the five prior work appears two times in the study. We show the results of this study in Figure \ref{fig:userstudy}. As seen, depending on the comparison baseline, between 87\%  to 100\% of users prefer our enhanced videos over prior work.

\subsection{Ablation Study}\label{subsec:ablation}

\begin{figure*}
     \centering
     \begin{subfigure}[b]{0.23\textwidth}
         \centering
         \includegraphics[width=\textwidth]{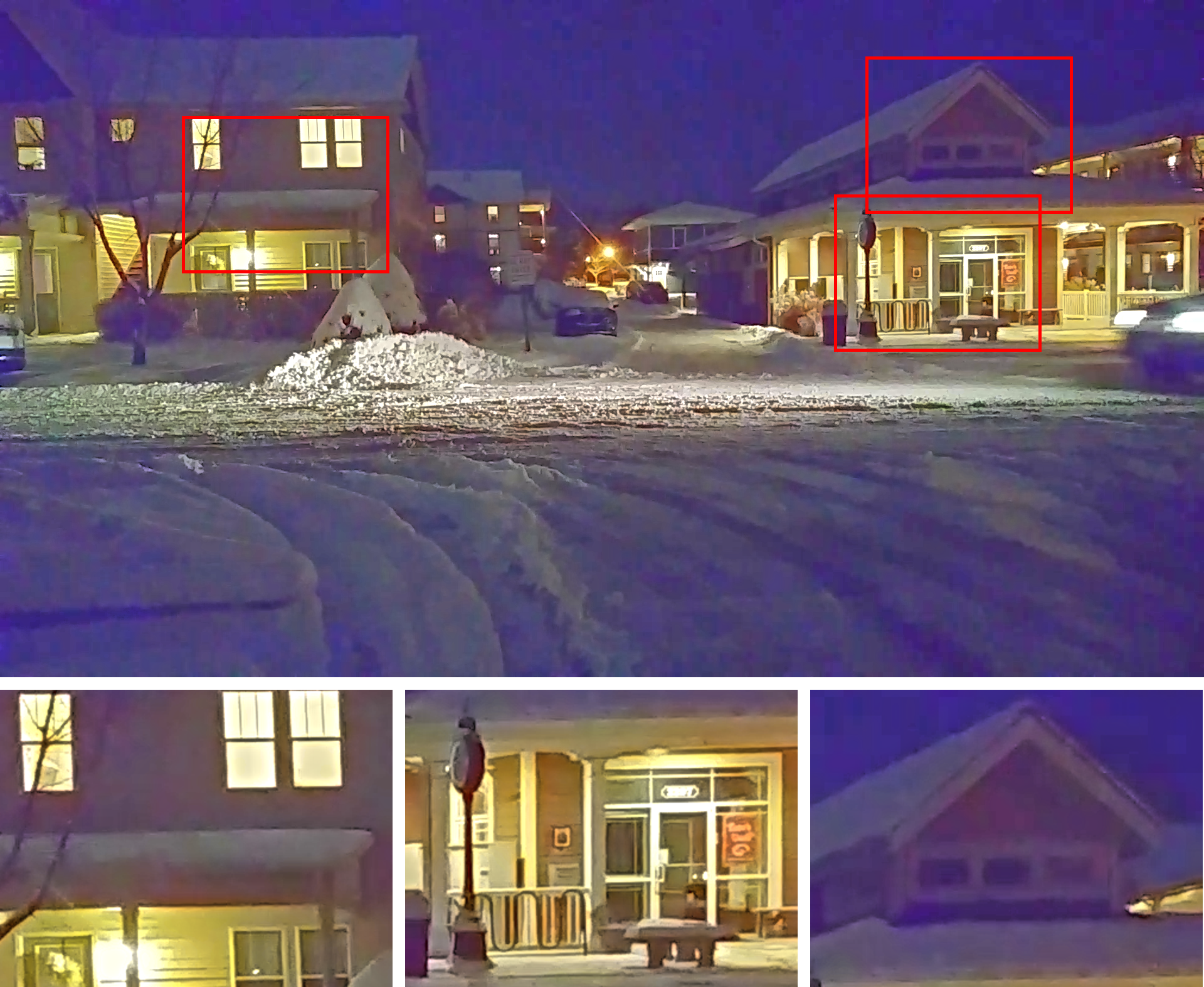}
         \caption{$\alpha=0.0$, $\beta=3$, denoising enabled.}
     \end{subfigure}
     \begin{subfigure}[b]{0.23\textwidth}
         \centering
         \includegraphics[width=\textwidth]{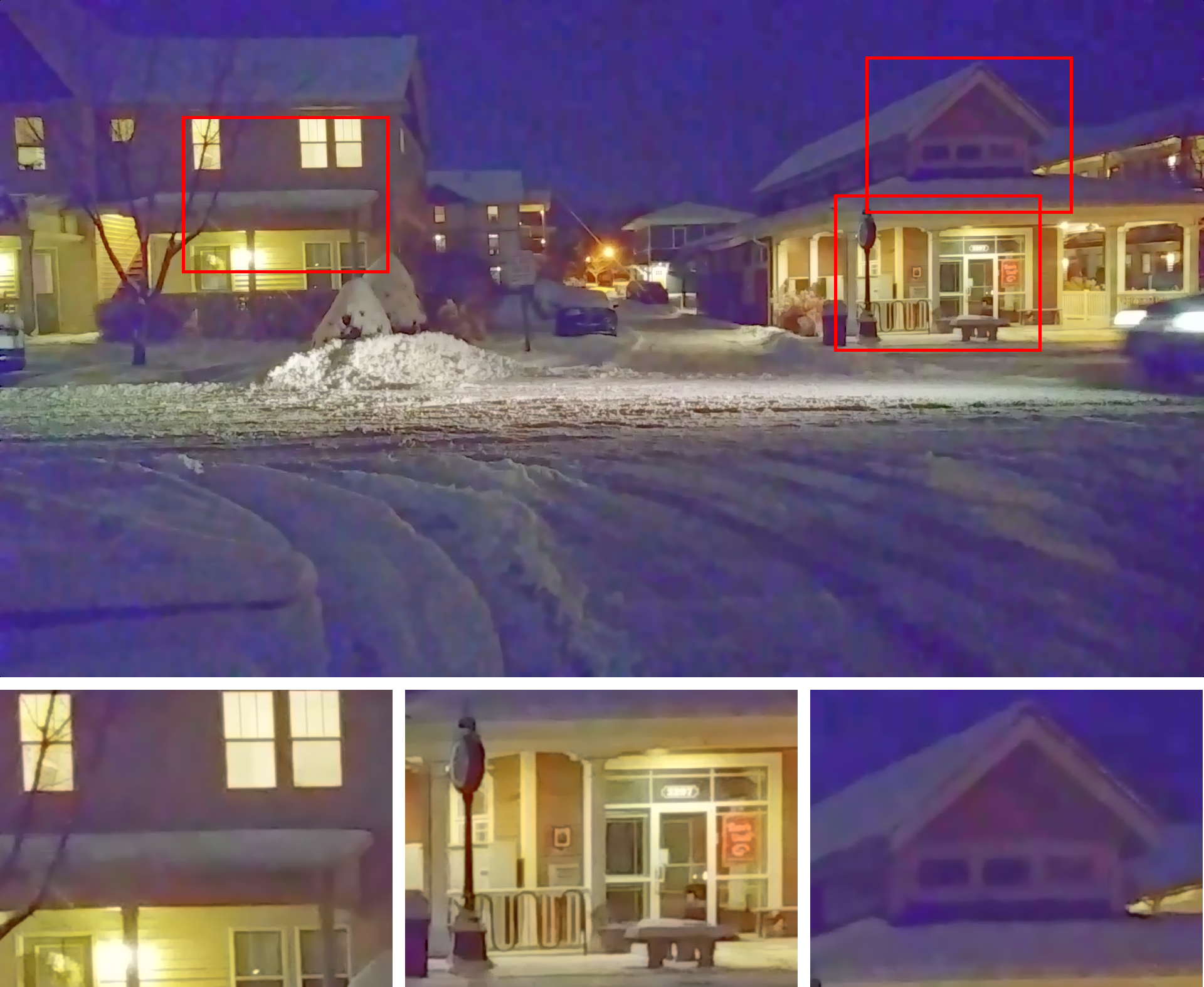}
         \caption{$\alpha=0.015$, $\beta=0$, denoising enabled.}
     \end{subfigure}
     \begin{subfigure}[b]{0.23\textwidth}
         \centering
         \includegraphics[width=\textwidth]{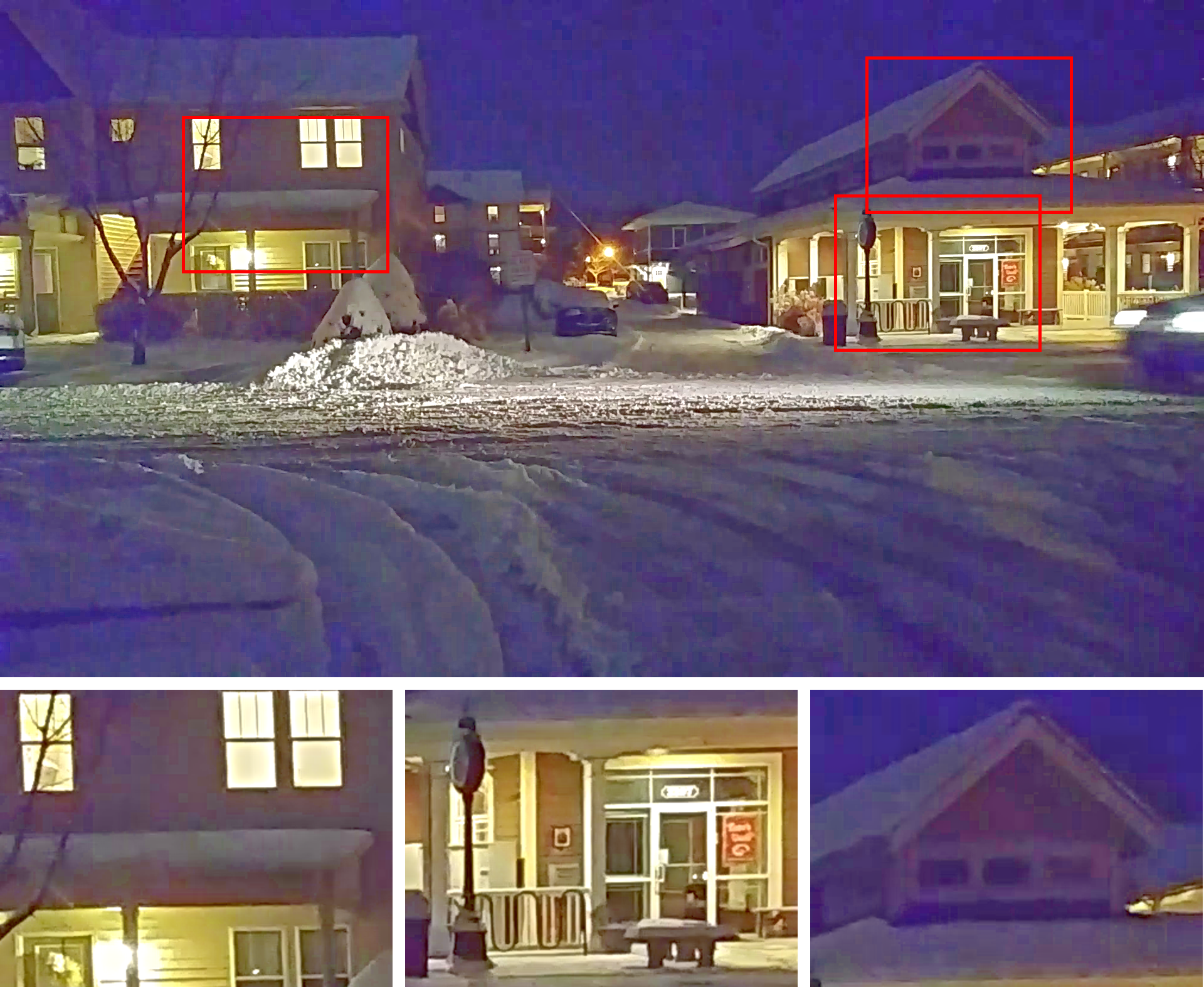}
         \caption{$\alpha=0.015$, $\beta=3$, denoising disabled.}
     \end{subfigure}
     \begin{subfigure}[b]{0.23\textwidth}
         \centering
         \includegraphics[width=\textwidth]{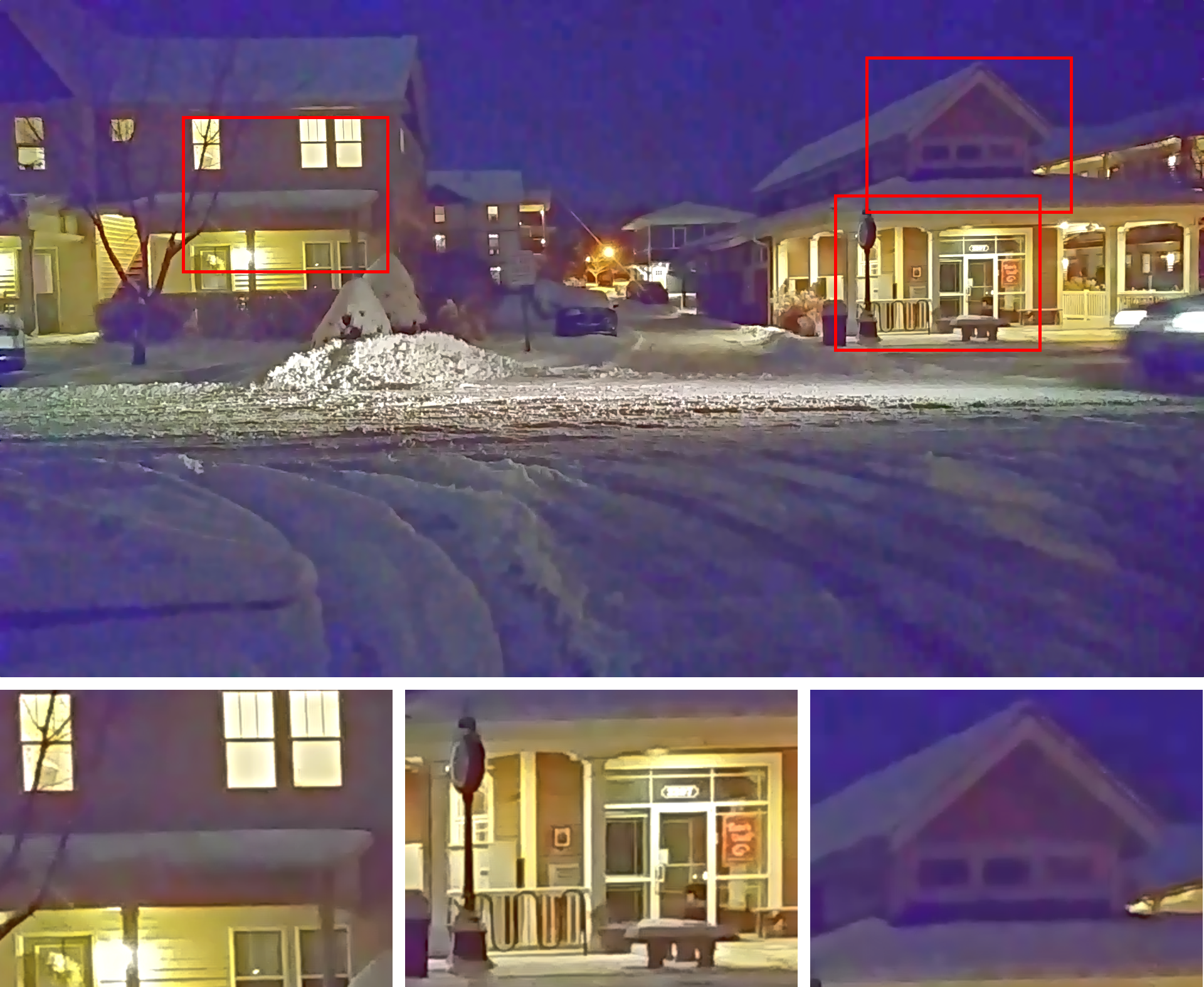}
         \caption{$\alpha=0.015$, $\beta=3$, denoising enabled.}
     \end{subfigure}
    \caption{Effect of parameters $\alpha$ and $\beta$ as well as the denoising operation on the enhanced frame's quality. }
    \label{fig:ablation1}
\end{figure*}

In \cite{azizi2020noise}, there is ablation study which shows the effectiveness of $\alpha$, $\beta$ and denoising of $\widehat{R}$ on the final enhanced image. Here, we study the effect of these three parameters on the future enhanced frames in Figure \ref{fig:ablation1}. It can be seen that cancelling parameter $\alpha$ and/or disabling the denoising operation results in noisy textures. Setting parameter $\beta$ to $0$ makes the edges of objects blurry and degrades texture preservance quality of the method.

\begin{figure*}
     \centering
     \begin{subfigure}[b]{0.23\textwidth}
         \centering
         \includegraphics[width=\textwidth]{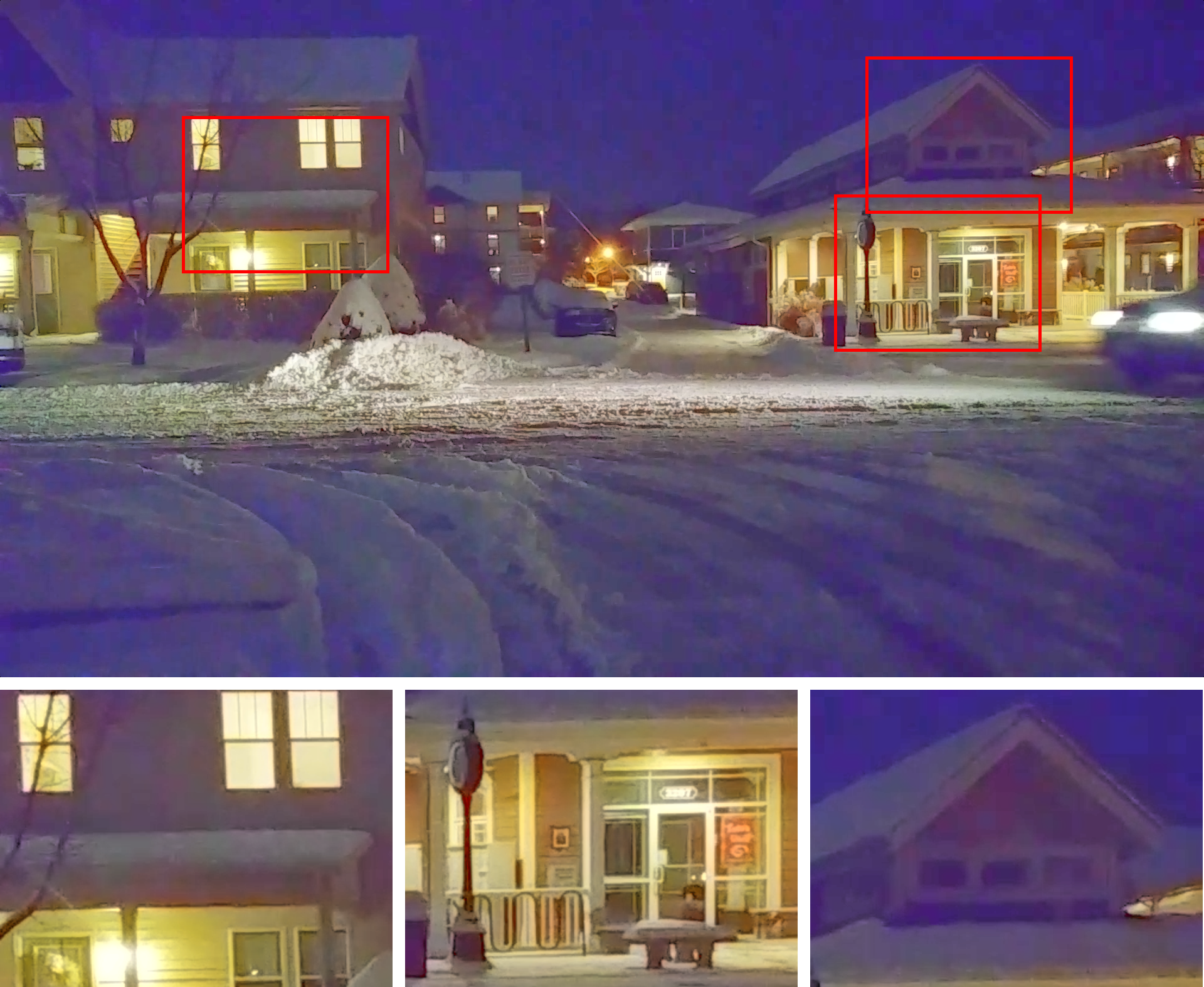}
         \caption{$1\times 1$ window}
     \end{subfigure}
     \begin{subfigure}[b]{0.23\textwidth}
         \centering
         \includegraphics[width=\textwidth]{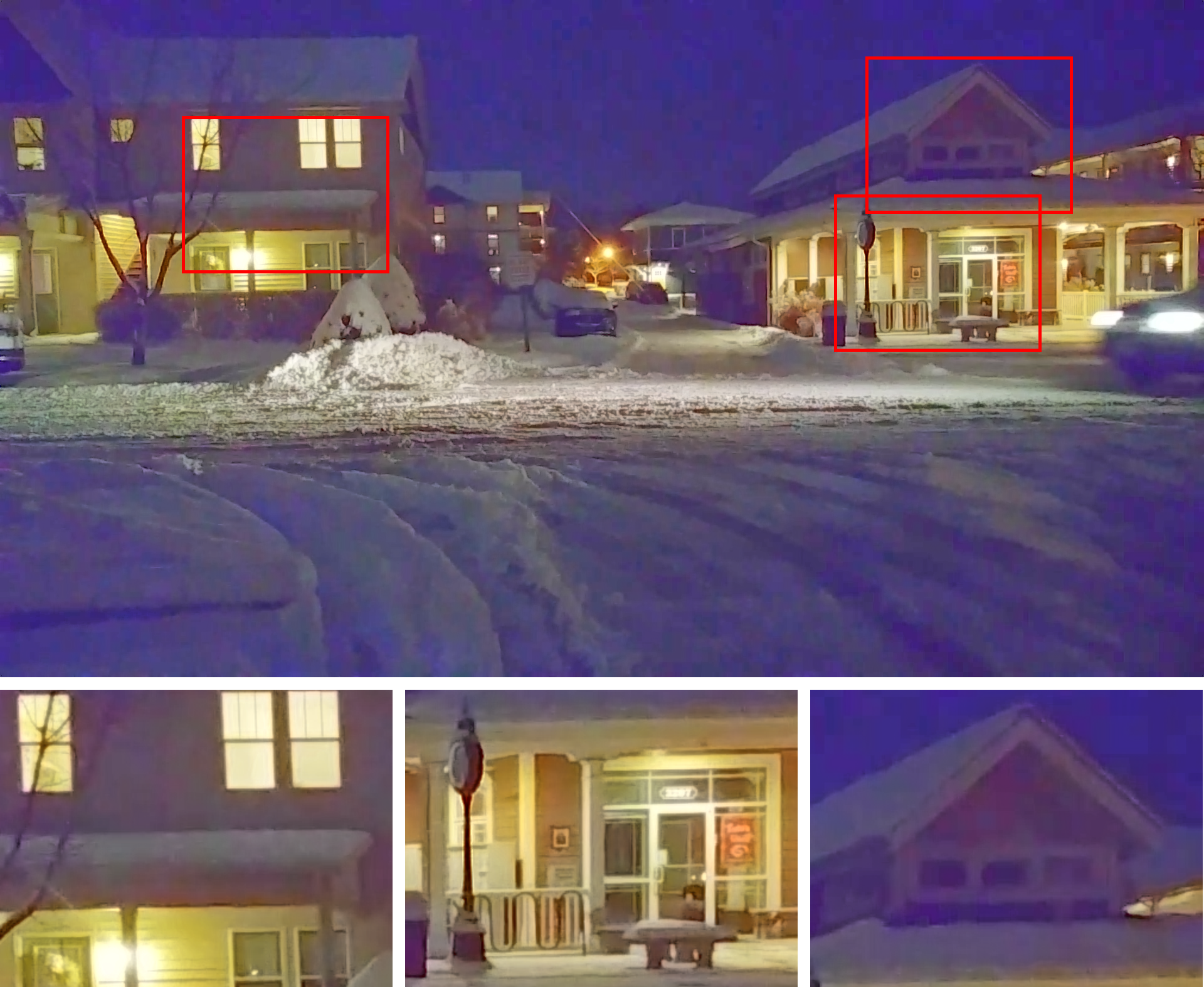}
         \caption{$3\times 3$ window}
     \end{subfigure}
     \begin{subfigure}[b]{0.23\textwidth}
         \centering
         \includegraphics[width=\textwidth]{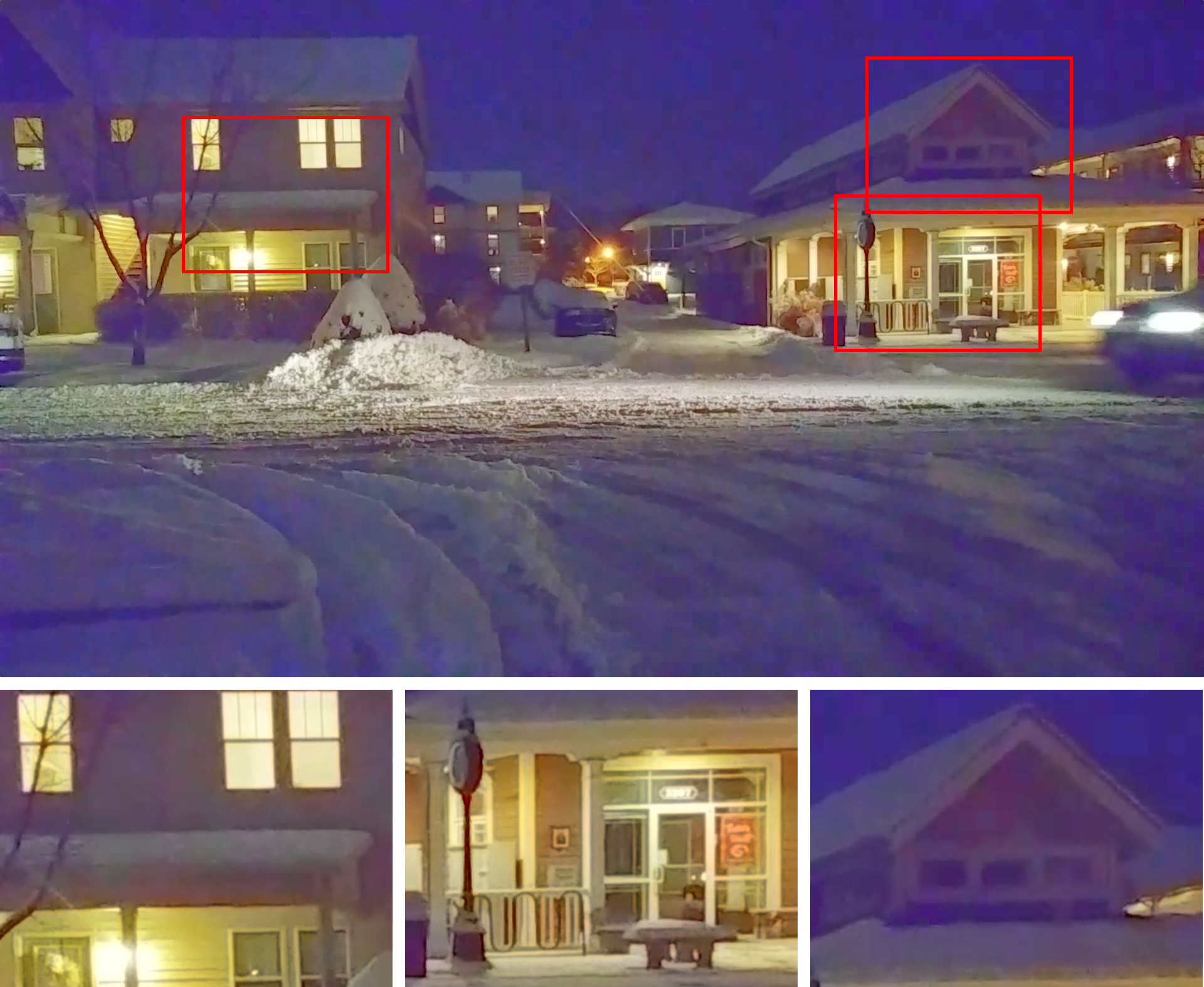}
         \caption{$5\times 5$ window}
     \end{subfigure}
     \begin{subfigure}[b]{0.23\textwidth}
         \centering
         \includegraphics[width=\textwidth]{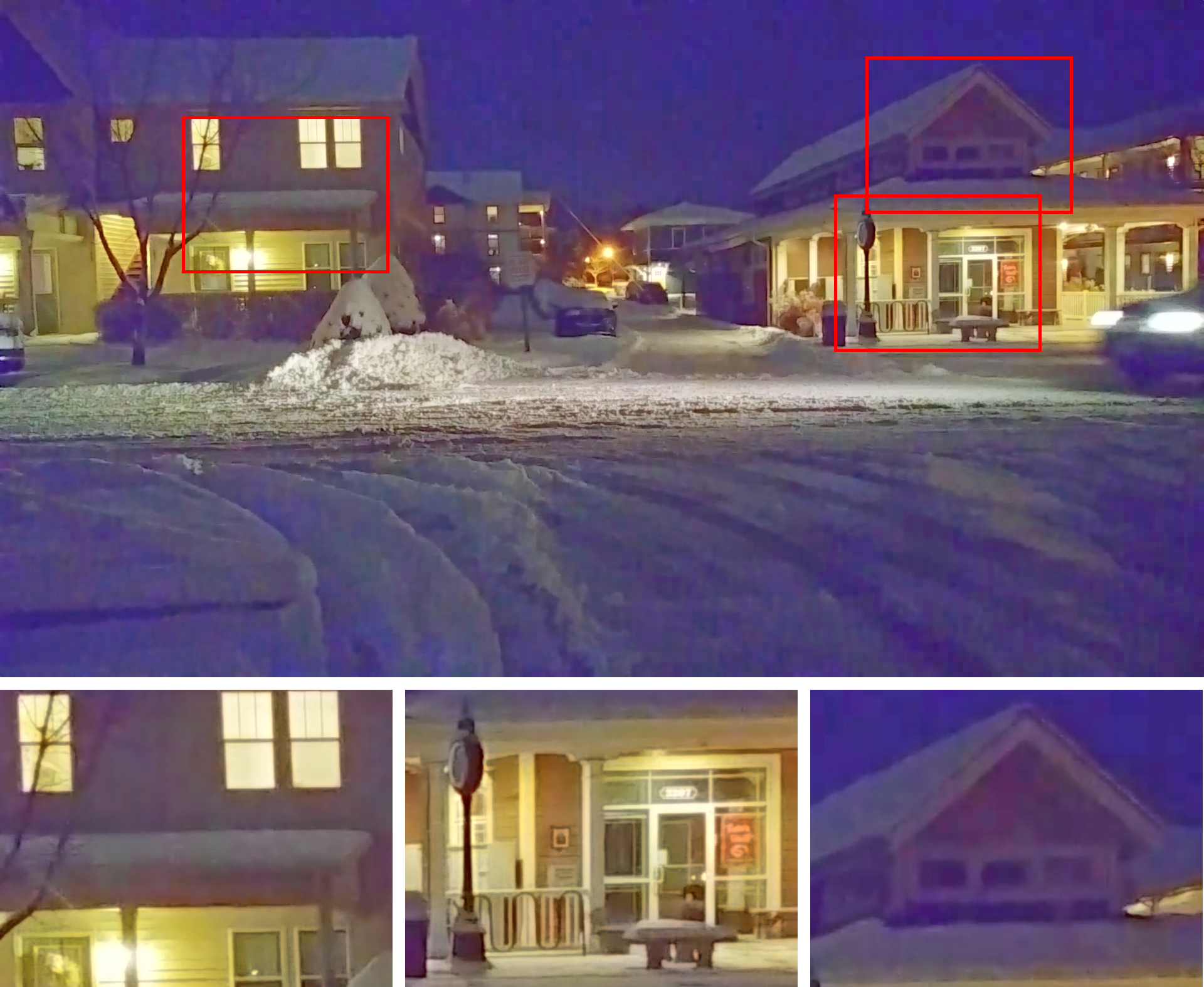}
         \caption{$7\times 7$ window}
     \end{subfigure}
    \caption{Effect of the regressor's window size on the quality of enhanced frames. }
    \label{fig:ablation2}
\end{figure*}

In section \ref{subsec:Video}, we explained that we selected the window size of Ridge regression to be 5. Here, we also analyze the effect of the window size on enhanced frames. Figure \ref{fig:ablation2} shows this analysis. A small window size causes artifacts to appear in the enhanced frame, e.g., some pixels on the street light become red instead of maintaining the black color.

\section{Conclusion}\label{sec:conclusion}

A new method for low-light video enhancement, called \method{}, was
proposed in this work.  The new self-supervised learning method is fully
adaptive to the test video. \method{} enhances a few
keyframes of the test video, learns a mapping from low-light to enhanced
keyframes, and finally uses the mapping to enhance the rest of the
frames. This approach enables \method{} to work without requiring
(paired) training data. Furthermore, we conducted a user study and
observed that participants preferred our enhanced videos in at least 87\%
of the tests.  Finally, we performed an ablation study to demonstrate
the contribution of each component of \method{}.




\section*{Acknowledgment}

This research was supported by a gift grant from Mediatek. The authors acknowledge the Center for Advanced Research Computing (CARC) at the University of Southern California for providing computing resources that have contributed to the research results reported within this publication. URL: \url{https://carc.usc.edu}.

\bibliographystyle{unsrt}  
\bibliography{references}  

\end{document}